\RequirePackage{fix-cm}
\documentclass[11pt]{article}
\usepackage{amsfonts,amstext,amssymb,amsmath}
\usepackage{bm,booktabs}
\usepackage{color,colordvi,comment}
\usepackage{enumitem}
\usepackage{graphicx}
\usepackage{hyperref}
\usepackage{mathptmx} 
\usepackage{outlines}
\usepackage{times}
\usepackage{xspace}

\newtheorem{definition}{Definition}
\newtheorem{proposition}{Proposition}
\newtheorem{theorem}{Theorem}
\newtheorem{lemma}{Lemma}

\newcommand{\bhat}[1]{\hat{\mathbf{#1}}}
\newcommand{\btil}[1]{\tilde{\mathbf{#1}}}

\newcommand{\mb}[1]{\mathbf{#1}}
\newcommand{\ubar}[1]{\underline{#1}}
\newcommand{\realvec}[1]{\mathbb{R}^{#1}}
\newcommand{\realmat}[2]{\mathbb{R}^{#1 \times #2}}
\newcommand{\upbrac}[1]{^{(#1)}}
\newcommand{\Hin}{H^{\mathrm{in}}}
\newcommand{\ubarHin}{\ubar{H}^{\mathrm{in}}}
\newcommand{\narrownet}{narrow-net\xspace}
\newcommand{\mediumnet}{medium-net\xspace}
\newcommand{\widenet}{wide-net\xspace}

\newcommand{\embpla}{\mathcal{P}_{\ubar{\theta}}}
\newcommand{\neusplit}{\mathtt{split}}
\newcommand{\Lam}[1]{\Lambda_{\mathrm{#1}}}

\newif\ifsubmit

\submittrue

\ifsubmit
\newcommand{\ruoyu}[1]{}
\newcommand{\dingtian}[1]{}
\newcommand{\dawei}[1]{}

\else
\newcommand{\ruoyu}[1]{{\textcolor{blue}{[Ruoyu: #1]}}}
\newcommand{\dingtian}[1]{{\textcolor{orange}{[Tian: #1]}}}
\newcommand{\dawei}[1]{{\textcolor{green}{[Dawei: #1]}}}

\fi
\begin{document}

\title{A Geometric Characterization of the Stationary Plateau for Two-Layer Neural Networks
}
%\subtitle{Do you have a subtitle?\\ If so, write it here}

%\titlerunning{Short form of title}        % if too long for running head

\author{Tian Ding\thanks{Shenzhen International Center of Industrial and Applied Mathematics, Shenzhen Research Institute of Big Data, China; Shenzhen Loop Area Institute, China; AutoKernel, China. Email: dingtian@sribd.cn. Part of the work was done when Tian Ding was visiting Ruoyu Sun in UIUC.},
\and Dawei Li\thanks{University of Minnesota Twin Cities. Email: li004678@umn.edu.},
\and Ruoyu Sun\thanks{School of Data Science, The Chinese University of Hong Kong, Shenzhen, China; Shenzhen International Center of Industrial and Applied Mathematics, Shenzhen Research Institute of Big Data, China; Shenzhen Loop Area Institute, China. Email: sunruoyu@cuhk.edu.cn.}}

%\authorrunning{Short form of author list} % if too long for running head

\date{\today}
% The correct dates will be entered by the editor
\maketitle

\begin{abstract}
We investigate the geometric structure of stationary plateaus that arise in the loss landscape of two-layer neural networks with smooth activation functions. We focus on the phenomenon of ``neuron splitting'' where duplicating a hidden neuron yields an affine set of stationary points in a wider network. We provide a comprehensive classification of all stationary points on these plateaus, determining under what conditions they constitute local minima or saddle points. Our characterization hinges on a per-neuron curvature object we term the ``inner Hessian'' matrix. Our analysis reveals that the definiteness of the inner Hessian and the choice of splitting coefficients jointly dictate the local geometry of the plateau. We show that ``splitting'' a local minimum can yield either a mixture of local minima and saddles or an all-saddle plateau, with a concrete sure-saddle region identified under mild assumptions. In contrast, splitting a saddle point always produces a plateau of saddle points. Our results unify and extend prior landscape analyses, elucidating when and how model expansion preserves or alters the nature of stationary points. These findings offer new geometric insights into the effects of width expansion and reparameterization in neural networks.
\end{abstract}

\section{Introduction}
Deep neural networks have achieved state-of-the-art performance across a wide range of applications, including natural language processing, computer vision, speech and audio processing, autonomous driving, and robotics.
A striking aspect of this progress is that training these models essentially amounts to solving highly nonconvex optimization problems that could, in principle, be riddled with poor stationary points. However, in practice, many neural networks are optimized efficiently and achieve excellent solutions at scale.
Despite these empirical advances, a complete theoretical understanding of this phenomenon remains elusive. From an optimization perspective, one of the key challenges arises from the complex geometric properties that characterize the loss surface \cite{oneto2023we}.

To tackle this challenge, a growing body of research has focused on the \emph{loss landscape} perspective, which examines the local geometry of stationary points and their relationship to optimization dynamics. 
Recent work has observed the existence of bad local minima in various regimes \cite{ding2022suboptimal, liu2021non}.
In the meantime, a number of benign geometric and topological features of the loss landscape of neural networks have also been uncovered, including the disappearance of basin/valley structures as network width increases \cite{venturi2018neural,li2018benefit}, as well as connectivity between global minima \cite{garipov2018loss, lin2024exploring}.
Collectively, these results suggest that, despite the nonconvexity and the possible presence of bad local minima, neural-network loss landscapes often possess global structures that are unexpectedly favorable to optimization. Such structures may help explain aspects of the success of gradient-based methods in practice.
However, this picture remains incomplete. In particular, it is still unclear how bad local minima arise, how they are organized together with other stationary points, and how they can coexist with the benign global structures observed in neural-network landscapes. 
This motivates a more detailed study of the local geometry of stationary points.

In this paper, we refine the existing picture by focusing on the \emph{geometry} of two–layer neural networks with smooth activation functions. Our central object is a special class of regions in parameter space that consist entirely of stationary points. We call these regions \emph{stationary plateaus}. Specifically, a stationary plateau is an affine set along which the network’s input–output function (and hence the empirical loss) remains invariant, while every point on the set is a stationary point of the empirical loss. Such plateaus arise naturally via \emph{neuron splitting} operation \cite{fukumizu2000local}, which embeds a narrower network into a wider one by duplicating a hidden neuron: incoming weights are kept unchanged, while the outgoing weight of the original neuron is redistributed among its duplicates \cite{zhang2021embedding, zhang2022embedding}. 
Throughout the paper, we refer to the neuron that is split in the original network as \emph{parent neuron} and the newly created neurons in the wider network as \emph{daughter neurons}. 
Prior work has observed that stationarity is preserved after neuron splitting: if a weight configuration is a stationary point, then its split counterparts form an affine stationary set in a wider network (termed a “plateau” in the literature) \cite{fukumizu2000local, fukumizu2019semi, zhang2021embedding, zhang2022embedding, simsek2021geometry}.
Furthermore, prior work has also taken an initial step in examining what types of stationary points appear on these plateaus, whether they are local minima or saddles under various conditions\cite{fukumizu2000local, zhang2022embedding}.

Neuron splitting offers a particularly clean lens on how increasing width reshapes the local geometry of the loss landscape. Starting from a stationary point $\underline{\theta}$ of a narrow network, splitting one hidden neuron into multiple duplicates produces an affine family of stationary points in the wider network: the \emph{stationary plateau} $\mathcal{P}_{\underline{\theta}}=\{\mathrm{split}(\underline{\theta},\lambda):\lambda\in\Lambda\}$. The dimension of this plateau grows with the number of duplicates (equivalently, with width), so widening does not merely create new isolated critical points—it \emph{unfolds} a single stationary point into a higher-dimensional stationary set along which the network function (and the loss) remains invariant.

\paragraph{Our contributions: a comprehensive classification of stationary points on split plateaus.} We undertake a systematic study of what happens to a given stationary point when the network is widened by splitting \emph{an arbitrary hidden neuron} into \emph{an arbitrary number of duplicates}. We provide a comprehensive characterization of the stationary plateau in the wider network: we determine whether each point on the plateau is a local minimum or a saddle, under transparently checkable conditions at the original (narrow-network) stationary point. Informally, our characterization can be summarized as follows.

\begin{itemize}
\item If the narrow-network stationary point is a (nondegenerate) local minimizer, then the split plateau in the wide network is either \emph{mixed} (containing both local minima and saddles) or \emph{purely saddle}; which case occurs is decided by a local geometric criterion at the parent neuron.
\item If the narrow-network stationary point is a saddle, then \emph{every} point on the split plateau is a saddle (i.e., saddles beget saddles).
\item In addition, assuming the narrow-network stationary point is not degenerate, we identify a concrete \emph{surely-saddle region} within the plateau: any coefficient choice in this region yields a saddle in the wide network.
\end{itemize}

The above results establish a \emph{plateau-type evolution law} under widening the neural network. Because neural networks admit no nontrivial local maxima (see Lemma \ref{lem:no_localmax}), stationary points are either local minima or saddles, and stationary plateaus naturally fall into three types: (I) \emph{all-localmin} plateaus (every point is a local minimum), (II) \emph{localmin--saddle} plateaus (local minima and saddles coexist), and (III) \emph{all-saddle} plateaus (every point is a saddle). Our results reveal a sharp and conceptually useful picture of how these types arise and evolve under width expansion: splitting a saddle point always yields an all-saddle plateau (\emph{saddles beget saddles}); in contrast, splitting a local minimum \emph{generically destroys} the all-localmin property. Except for degenerate situations (where the loss is locally insensitive to perturbing the split neuron), widening a local minimum produces a plateau that contains saddle points. In this sense, width expansion tends to convert an isolated ``potentially bad'' local-minimum region into a stationary set that contains escape directions.

Technically, the characterization hinges on a per-neuron curvature object: an \emph{inner Hessian matrix} associated with the parent neuron’s weights. The inner Hessian governs how small asymmetric perturbations of the daughter neurons' weights change the loss. The definiteness of this inner Hessian, together with the splitting coefficients, dictates whether a given plateau point is a local minimum or a saddle. (Formal definitions and the precise theorems are provided in Section~\ref{sect:main_results}.)

\section{Related Works}

\paragraph{Neuron splitting and stationary plateaus.}
It is widely recognized that stationary points of neural-network loss, including spurious local minima \cite{safran2018spurious, he2020piecewise, ding2022suboptimal, liu2021non} and degenerate saddle points \cite{dauphin2014identifying, achour2024loss}, are key geometric features of the loss landscape. These structures can substantially affect the training dynamics, slowing down learning \cite{dauphin2014identifying, safran2018spurious, du2017gradient}. The study of neuron splitting provides a conceptual foundation for understanding how neural networks reorganize their internal structure and give rise to stationary plateaus in the loss landscape. 
A number of works have examined the stationary points created under neuron splitting operations. 
Most of these analyses \cite{fukumizu2000local, fukumizu2019semi, simsek2021geometry, zhang2021embedding, zhang2022embedding, zhang2024geometry} focus on sufficient conditions under which saddles may occur in the stationary plateau, while \cite{wu2024loss} provides a comprehensive classification of such plateaus for ReLU-like networks. 
In contrast to the works that certify the existence of saddle points after embedding, our results classify the coefficient regions on the split plateau in which local minimality is preserved or destroyed for networks with smooth activation functions by introducing the tool of inner-Hessian analysis. 

\paragraph{Geometric structure of stationary points.}
Besides neuron splitting, several lines of works investigate the geometric structures of stationary points in neural loss landscapes from multiple perspectives. 
These include the symmetry-induced connectivity of stationary points \cite{brea2019weight, simsek2021geometry}, the connectivity of global minima \cite{garipov2018loss, lin2024exploring}, the manifold characterization of global-minimum structures \cite{cooper2021global}, and the emergence of saddle branches \cite{zhang2024geometry}. 
By analyzing neuron splitting, our work sharpens these geometric pictures through the viewpoint of split plateaus, which provides a unified mechanism for understanding the creation of minima, the formation of the stationary plateau, and the emergence of saddle or bifurcation structures. 

\paragraph{Wide-network geometry and training dynamics.}
Neuron splitting can also be viewed as a mapping from narrow to wide neural networks, or more generally as a structured reparameterization of the network landscape \cite{levin2025effect}. 
Specifically, neuron splitting provides a local geometric mechanism by which increasing width can reshape bad regions of a narrow-network landscape into more escapable structures. 
This viewpoint is consistent with several broader explanations of why sufficiently wide networks often exhibit more benign optimization behavior, such as the smooth landscape \cite{sagun2017empirical, li2018visualizing}, the disappearance of spurious local minima or basins \cite{kawaguchi2016deep, nguyen2017loss2, laurent2018deep, li2018benefit}, benign training dynamics or fast convergence in GD \cite{du2018gradienta, du2018gradientb, allen2019convergence, zou2019improved, xu2023linear}.

\paragraph{Broader perspectives and connections.}
Our conclusions are consistent with—and help structurally explain—several observations in optimization dynamics and architecture-growth algorithms. 
The local-min–saddle line structure derived in our analysis provides a local geometric analogue of the ``river-valley landscape'' view of the training dynamics, where the trajectory moves along a low-curvature progress direction while oscillating across other directions \cite{wen2024understanding}. 
It is also aligned with splitting-based architecture growth methods, such as the ``split-and-descent'' algorithm \cite{wu2019splitting}, where splitting a neuron reveals new descent directions for escaping stationary regions.

Beyond theory, these findings imply practical guidance for \emph{model expansion} and width-adaptation procedures—splitting around locally effective neurons tends to generate saddles unless specific coefficient patterns are chosen, while splitting around locally flat or constant neurons can preserve local minimality. 
Such insights may inform principled widening or neural architecture growth strategies.

\section{Preliminaries}
\label{sect:prelim}
\subsection{Network Settings}
\label{sect:network_settings}
We consider a scalar-output one-hidden-layer network given by
\begin{equation}
\label{eq:net_m}
f\left(\mb{x};\theta\right) = \mb{v}^\top \sigma\left(W\mb{x}\right)
\end{equation}
where $\mathbf{x}\in \realvec{d}$ represents the input data, $W\in \realmat{m}{d}$ is the hidden weight, $\mb{v}\in \realvec{m}$ is the output weight, $\theta = (\mb{v},W)$ denotes the configuration of the network weights, and $\sigma: \mathbb{R} \rightarrow \mathbb{R}$ is an activation function applied component-wise to the input. The activation function is assumed to be twice continuously differentiable.

Given $n$ training input samples $\{\mb{x}^{(n)}\}^n_{k=1}$ and corresponding labels $\{y^{(n)}\}^n_{k=1}$, we consider the quadratic loss function, i.e.,
\begin{equation}
\label{eq:loss_origin}
L(\theta) \triangleq \frac{1}{2}\sum^n_{k=1}\left( f(\mb{x}^{(k)};\theta) - y^{(k)}\right)^2.
\end{equation}
which is often called the empirical loss or the empirical risk
in machine learning.
To recast the problem in a more compact form, let the matrix $X = [\mb{x}^{(1)}, \mb{x}^{(2)}, \cdots, \mb{x}^{(n)}]\in \realmat{d}{n}$ and the vector $\mb{y} = [y^{(1)}, y^{(2)},..., y^{(n)}]^\top \in \realmat{n}{1}$ to collect all the training inputs and training outputs, and define the prediction vector 
corresponding to the weight $\theta$ as
\begin{equation}
    \bhat{y} =  \left[f(\mb{x}^{(1)};\theta), f(\mb{x}^{(2)};\theta), \cdots, f(\mb{x}^{(n)};\theta)\right]^\top \in \realmat{n}{1}.
\end{equation}
Then, the empirical loss \eqref{eq:loss_origin} can be reformulated as
\begin{equation}
\label{eq:loss}
    L(\theta) = \frac{1}{2}\left\|\bhat{y}-\mb{y}\right\|^2_2.
\end{equation}
The training of the neural network aims to find a $\theta^*$ in the weight space $\Theta\triangleq \realvec{m} \times \realmat{m}{d}$ that minimizes \eqref{eq:loss}. 

Before we proceed, we introduce some important notations. We define $\mb{z}_i \in \realvec{n}$ to be the output vector of the $i$-th hidden neuron, given by
\begin{equation}
    \mb{z}_i^\top \triangleq \sigma(\mb{w}_i^\top X), \quad i\in[m]
\end{equation}
where $\mb{w}_i^\top \in \realvec{1\times d}$ is the $i$-th row of $W$, namely, the input weights to the $i$-th hidden neuron. We also define the first-order and second-order derivatives, denoted by $\mb{z}'_i, \mb{z}''_i \in \realvec{n}$ respectively, as
\begin{equation}
    {\mb{z}_i'}^\top \triangleq \sigma'(\mb{w}_i^\top X), \quad {\mb{z}_i''}^\top \triangleq \sigma''(\mb{w}_i^\top X).
\end{equation}
Note that the vectors $\mb{z}_i, \mb{z}'_i, \mb{z}''_i$ only involve the hidden weight $\mb{w}_i$, and hence are independent for different hidden neurons. These vectors will play an important roles in our analysis.

\subsection{Neuron Splitting}
This paper focuses on the optimization landscape of the empirical loss function \eqref{eq:loss}. Specifically, we study the stationary points of \eqref{eq:loss} via a hierarchical mapping termed ``neuron splitting''. Neuron splitting is a mapping from the weight space of a \emph{narrow} network to that of a \emph{wide} network. To our knowledge, it was first discovered in reference \cite{fukumizu2000local}. 
By ``splitting'' the hidden neurons, a stationary point of a single-hidden-layer network gives rise to a set of stationary points of a wider network, constituting an affine space in the weight space of the wider network. We characterize this hierarchical relationship by establishing sufficient and/or necessary conditions on whether the stationary points of the wider network to be local minima or saddle points.

\begin{figure}
\centering
\includegraphics[width=0.75\textwidth]{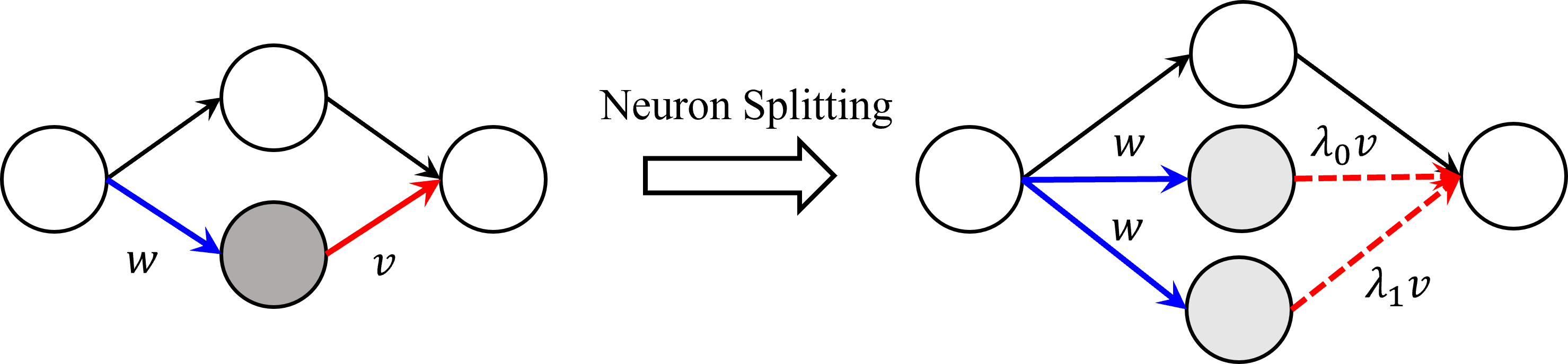}
\caption{An illustration of neuron splitting. One hidden neuron of the width-2 network is split into two, yielding a width-3 network. The hidden weight $\mb{w}$ is preserved, while the output weight spreads over the new neurons according to coefficients $\lambda_0$, $\lambda_1$ with $\lambda_0 + \lambda_1 = 1$.}
\label{fig:neuron_splitting}
\end{figure}

To begin with, we formally introduce neuron splitting. Consider a narrow single-hidden layer network with $r$ neurons and a wide single-hidden layer network with $m$ neurons. For simplicity, we refer to the narrow and the wide networks as \emph{\narrownet} and \emph{\widenet}, respectively. We assume that \narrownet and \widenet share the same activation function $\sigma$ and the same training data $(X,\mb{y})$.
For \widenet, we directly adopt the notations introduced in Section \ref{sect:network_settings}. For \narrownet, we use $\ubar{\theta} = (\ubar{\mb{v}},\ubar{W})$ to denote the weight configuration, where $\ubar{\mb{v}}\in \realvec{r}$, $\ubar{W}\in \realmat{r}{d}$. Then, the prediction of 
the \narrownet is represented by
\begin{equation}
    \label{eq:net_r}
    \ubar{f}(\mb{x};\ubar{\theta}) =
    \ubar{\mb{v}}^\top \sigma(\ubar{W}\mb{x}).
\end{equation}
The prediction vector $\ubar{\bhat{y}}$ and the empirical loss function $\ubar{L}(\ubar{\theta})$ of \narrownet are defined in the same way as \widenet. Further, in this paper, we always use underlined notations to denote variables associated with the \narrownet (except the shared training data and the activation function). 

We say that a neuron of a narrow network is ``split'' if it is replaced by several new neurons (entailing a wider network), with hidden weights unchanged and the output weight spreading over the new neurons. Formally, we have the following definition.
\begin{definition}[Neuron splitting]
\label{def:neuron_splitting}
Let $\theta=(\mb{v},W)$ and $\ubar{\theta}=(\ubar{\mb{v}}, \ubar{W})$ be weight configurations of \widenet and \narrownet, respectively, and $\pmb{\lambda} =[\lambda_0, \lambda_1, \cdots, \lambda_{m-r+1}]$ be an $(m-r+1)$-dimensional vector\footnote{To ease the notation in the analysis, we let the entry index of $\pmb{\lambda}$ start from $0$.}. We say that $\theta$ is generated by splitting the $r$-th neuron of \narrownet at $\ubar{\theta}$ with coefficients $\pmb{\lambda}$ if 
\begin{subequations}
\label{eq:def_splitting}
\begin{align}
\label{eq:def_splitting_a}
        \mb{w}_i &= \ubar{\mb{w}}_i,\quad v_i=\ubar{v}_i,\quad  \forall i \in [r-1] \\
\label{eq:def_splitting_b}
    \mb{w}_{i'} &= \ubar{\mb{w}}_r, \quad \forall i' \in [r:m] \\
\label{eq:def_splitting_c}
    v_{i'} &= \lambda_{i'-r} \ubar{v}_r,\quad  \forall i'\in[r:m].    
\end{align}
\end{subequations}
Let $\Lambda\triangleq\{\pmb{\lambda}\in \realvec{m-r+1}: \sum^{m-r}_{j=0}\lambda_j = 1\}$. We define the neuron splitting operator as
\begin{equation}
\neusplit:\ubar{\Theta} \times \Lambda \rightarrow \Theta    
\end{equation}
such that $\theta= \neusplit(\ubar{\theta},\pmb{\lambda})$ is given by \eqref{eq:def_splitting} for any $\ubar{\theta}\in \ubar{\Theta}$ and $\pmb{\lambda} \in \Lambda$.
\end{definition}

For simplicity, we term the $r$-th neuron of \narrownet as the \emph{parent neuron} and the $r$-th to the $m$-th neurons of \widenet as the \emph{daughter neurons}. From \eqref{eq:def_splitting_a}, we see that the weights associated with the first $r-1$ neurons are unchanged during the splitting. \eqref{eq:def_splitting_b} means that the hidden weight to the parent neuron is copied to all the daughter neurons, and \eqref{eq:def_splitting_c} means that the output weight of the parent neuron spreads over the daughter neurons according to coefficients $\{\lambda_{j}\}^{m-r}_{j=0}$. An example of neuron splitting is illustrated by Fig. \ref{fig:neuron_splitting}. We note that splitting the last neuron of \narrownet is only for the convenience of notation. Definition \ref{def:neuron_splitting} and all the results in this paper directly apply to the case of splitting any other neuron. 

An important property of neuron splitting is that it preserves the input-output mapping and the empirical loss of \narrownet. Moreover, it maps a stationary point of \narrownet to a stationary point of \widenet. We term this property ``stationarity preservation''. Formally, we have the following proposition. 
\begin{proposition}
\label{prop:stat_preserve}
Let $\ubar{\theta}$ be a weight configuration of \narrownet, and $\theta = \neusplit(\ubar{\theta}, \pmb{\lambda})$ for an arbitrary $\pmb{\lambda}\in \Lambda$. We have
\begin{enumerate}
    \item $f(\mb{x}; \theta) = \ubar{f}(\mb{x}; \ubar{\theta})$ and $L(\theta) = \ubar{L}(\ubar{\theta})$, $\forall \mb{x}\in \realvec{d}$;
    \item $\theta$ is a stationary point of $L$ if and only if $\ubar{\theta}$ is a stationary point of $\ubar{L}$.
\end{enumerate}
\end{proposition}
Stationarity preservation was first discovered in \cite{fukumizu2000local} (Theorem 1), and was then exploited by \cite{fukumizu2019semi,zhang2021embedding} to develop their theorems. Note that all the above prior works only involved the ``if'' part. That is, a stationary point of \narrownet gives rise to stationary points of \widenet. We supplement the property by the ``only if'' part. For completeness, we provide a full proof of Proposition \ref{prop:stat_preserve}, which is deferred to Appendix \ref{app:pfPropSP}.

From Proposition \ref{prop:stat_preserve}, we see that a stationary point $\ubar{\theta}$ of \narrownet gives rise to a set of stationary points 
\begin{equation}
    \mathcal{P}_{\ubar{\theta}} \triangleq \{\theta = \neusplit(\ubar{\theta}, \pmb{\lambda}): \pmb{\lambda} \in \Lambda\}
\end{equation}
of \widenet, all with the same empirical loss. Because $\neusplit$ is an affine mapping with respect to $\pmb{\lambda}$, $\mathcal{P}_{\ubar{\theta}}$ defines an affine subspace in $\Theta$, which can be seen as an affine plateau embedded in the loss landscape of \widenet. Throughout this paper, we refer to $\ubar{\theta}$ as the \emph{original point}, $\theta = \neusplit(\ubar{\theta}, \pmb{\lambda})$ as the \emph{embedded point} of neuron splitting, and $\embpla$ as the \emph{stationary plateau} or \emph{embedded plateau}. Most of this paper's effort will focus on characterizing the types of embedded points, i.e., whether they are local minima or saddle points, under given conditions on the original points.

\section{Main Results}
\label{sect:main_results}
By neuron splitting, a stationary point of \narrownet gives rise to a stationary plateau. A stationary point of a function could be a local maximum, local minimum or a saddle point. However, it can be shown that for neural networks, there is no non-trivial local maximum.
\begin{lemma}
\label{lem:no_localmax}
Consider a single-hidden-layer network with empirical loss $L$ given by \eqref{eq:loss}. A weight configuration $\theta \in \Theta $ is a local maximum of $L$ if and only if $L$ is locally constant at $\theta$.
\end{lemma}
The proof of Lemma \ref{lem:no_localmax} is deferred to Appendix \ref{app:pfLemNL}. We note that although the lemma considers single-hidden-layer network with quadratic loss function, the conclusion can be generalized to neural networks of any width, any depth and any convex loss function, as discussed in Appendix \ref{app:pfLemNL}.

Notice that a locally-constant point can be seen as either a local-min or a local-max. Given Lemma \ref{lem:no_localmax}, we only need to consider two types of stationary points on the loss landscape: local minima and saddle points. Accordingly, a stationary plateau can be categorized into three classes:
\begin{enumerate}
    \item \emph{All-localmin plateau} consisting of local-mins only;
    \item \emph{Localmin-saddle plateau} consisting of both local-mins and saddles;
    \item \emph{All-saddle plateau} consisting of saddles only.
\end{enumerate}

\subsection{Neuron Splitting at Local Minimum}
We first study the embedded plateau when the original point of \narrownet is a local minimum. We show that the types of the embedded points are largely related to a matrix associated with the parent neuron, termed ``inner Hessian matrix''.

For the $i$-th neuron of \widenet, we define $\Hin_i \in \realmat{d}{d}$ to be the inner Hessian matrix whose entries are given by
\begin{equation}
\label{eq:def_inner_Hessian}
    (\Hin_i)_{j,j'} = v_i\langle \bhat{y} -\mb{y}, \mb{z}_i'' \circ \mb{x}_j \circ \mb{x}_{j'} \rangle,\quad j,j' \in [d], \quad i \in [m]
\end{equation}
where $\mb{x}_{j}, \mb{x}_{j'} \in \realvec{n}$ denote (the transposes of) the $j$-th and $j'$-th rows of input data matrix $X$. The inner Hessian matrix $\ubarHin_i$ for \narrownet is defined similarly as \eqref{eq:def_inner_Hessian}, with $v_i$, $\bhat{y}$, and $\mb{z}_i''$ replaced by their underlined version. Note that the inner Hessian matrix is a real symmetric matrix, and thus has real eigenvalues. The name ``inner Hessian'' arises from a decomposition of a partial Hessian matrix. Specifically, the partial Hessian of the empirical loss $L$ with respect to $\mb{w}_i$ can be decomposed as
\begin{equation}
\label{eq:partial_Hessian_decompose}
    H_L(\mb{w}_i) = J_{\bhat{y}}^\top(\mb{w}_i)J_{\bhat{y}}(\mb{w}_i) + \sum^n_{k=1} (\hat{y}_k-y_k) H_{\hat{y}_k}(\mb{w}_i)
\end{equation}
where $J_{\bhat{y}}(\mb{w}_i)\in \realmat{n}{d}$ is the Jacobian matrix of $\bhat{y}$ with respect to $\mb{w}_i$ and $H_{\hat{y}_k}(\mb{w}_i) \in \realmat{d}{d}$ is the Hessian matrix of $\hat{y}_k$ with respect to $\mb{w}_i$. It can be verified that the inner Hessian $\Hin_i$ is exactly the second term of the decomposition \eqref{eq:partial_Hessian_decompose}. Please see Appendix \ref{app:inner_hessian} for details.

Given that the original point is a local minimum of \narrownet, our first theorem establishes the connection between the embedded plateau and the inner Hessian of the parent neuron. 
\begin{theorem}
\label{thm:splitting_localmin}
Suppose $\ubar{\theta}$ is a local minimum of \narrownet with $\ubar{L}(\ubar{\theta})\not=0$. Denote
\begin{subequations}
\label{eq:def_lambda_pd_nd}
\begin{gather}
    \Lam{PD} \triangleq \{\pmb{\lambda} \in \Lambda: \lambda_j >0 , ~\forall j\in [0:m-r]\}\\
    \Lam{ND} \triangleq \{\pmb{\lambda} \in \Lambda: \exists j \in [0:m-r], \text{ s.t. } \lambda_j >0 , ~\lambda_{j'}<0, ~\forall j'\not=j\}.
\end{gather}
\end{subequations}
Let $\theta = \neusplit(\ubar{\theta}, \pmb{\lambda})$ be the embedded point of the \widenet by neuron splitting. Then
\begin{enumerate}
\item If $\ubarHin_r$ is positive definite (PD), then $\theta$ is a local minimum of \widenet for any $\pmb{\lambda}\in \Lambda_{\mathrm{PD}}$, and is a saddle point of \widenet for any $\pmb{\lambda}\in \Lambda\backslash\Lambda_{\mathrm{PD}}$;
\item If $\ubarHin_r$ is negative definite (ND), then $\theta$ is a local minimum of \widenet for any $\pmb{\lambda}\in \Lambda_{\mathrm{ND}}$, and is a saddle point of \widenet for any $\pmb{\lambda}\in \Lambda\backslash\Lambda_{\mathrm{ND}}$;
\item If $\ubarHin_r$ is indefinite (ID), then $\theta$ is a saddle point of \widenet for any $\pmb{\lambda}\in \Lambda$;
%\item For any $\ubarHin_r \not= \mb{0}$, $\theta$ is a saddle point for any $\pmb{\lambda}\in \Lambda_{\mathrm{saddle}}\triangleq \Lambda-\Lambda_{\mathrm{PD}}\cap\Lambda_{\mathrm{ND}}$.
\end{enumerate}
\end{theorem}

Theorem \ref{thm:splitting_localmin} shows that splitting a local-min of \narrownet can give rise to either a local-min saddle plateau or an all-saddle plateau. For PD, ND, or ID inner Hessian of the parent neuron, we provide a full characterization of the embedded plateau, i.e., identification of every point on the embedded plateau. 

However, given Theorem \ref{thm:splitting_localmin} alone, it is still not clear whether an all-localmin plateau can appear by splitting a local minimum. Specifically, if the inner Hessian is positive semi-definite or negative semi-definite (both having at least one zero eigenvalue), what can we tell about the embedded plateau? Our next result shows that, as long as the inner Hessian is \emph{non-zero}, i.e., having at least one non-zero entry, we can identify a region on the plateau that consists of only saddle points.

\begin{theorem}
\label{thm:surely_saddle}
Suppose that $\ubar{\theta}$ is a local minimum of \narrownet with $\ubar{L}(\ubar{\theta})\not=0$, and $\theta = \neusplit(\ubar{\theta}, \pmb{\lambda})$ is the embedded point of the \widenet by neuron splitting. Denote $\Lam{saddle} \triangleq \Lambda \backslash(\Lam{PD}\cup \Lam{ND})$ where $\Lam{PD}, \Lam{ND}$ are defined in \eqref{eq:def_lambda_pd_nd}. If $\ubarHin_r \not= \mb{0}$, then $\theta$ is a saddle point of \widenet for all $\pmb{\lambda}\in \Lam{saddle}$.
\end{theorem}

We notice that the region specified in Theorem \ref{thm:surely_saddle} is a positive-measure subset of the parameter space $\Lambda$. Identifying this ``surely-saddle region'' may potentially enable us to create computationally-tractable saddle points during the neuron splitting operation.

Next, we proceed even further to the rare case where the inner-Hessian is \emph{zero}. We show that excluding a degenerate case, splitting a local-min of \narrownet always entails saddle points on the embedded plateau. We first define two notions, termed ``locally effective neuron'' and ``locally constant neuron''. 

\begin{definition}[Locally Effective Neuron]
\label{def:effective_neruon}
The $i$-th neuron of a single-hidden-layer neural network is said to be locally effective at weight configuration $\theta$ if for any $\delta>0$, there exists $\mb{w}_i' \in B(\mb{w}_i,\delta)$ such that 
\begin{equation}
\left \langle \bhat{y}- \mb{y}, \sigma^\top(\mb{w}'^{\top}_i X)\right \rangle \not= 0.
\end{equation}
\end{definition}

\begin{comment}
【现有的组织】
Sect 3.2： Splitting local-min:
Theorem 1: inner Hessian PD, ND, or ID, characterizes all points.

Theorem 2: if not locally static (locally effective), then exists saddle. if locally constant, all points are local-min.

Theorem 3: if non-zero inner Hessian, we find a non-zero measure surely-saddle region. (points in other regions are unknown.)

Yaoyu: under some conditions, the split plateau surely has a saddle region (zero-measure)

Sect 3.3: 
Theorem 4: saddle → saddle

Splitting local-min:

1. Locally constant
2. Locally static
3. Zero inner Hessian (all-zero)
4. Zero eigenvalue exists inner Hessian
-- If 1, then 2. If 2, then 3. If 3, then 4.

full characterization
- If not 4, we can characterize all points. (we actually get a stronger conclusion.)
- If 1, we prove the non-existence of saddle points.

partial characterization (identify existence of saddles)
- If not 3, we can characterize a surely saddle region (non-zero measure set on the plateau).
- If not 2, we prove the existence of saddle points (a zero-measure set of points on the plateau).

What we don't know: 2 holds but not 1.

Splitting saddle:
All points on the plateau are saddles.
(Yaoyu: strict saddles → strict saddles)
\end{comment}

\begin{definition}[Locally Constant Neuron]
\label{def:constant_neruon}
The $i$-th neuron of a single-hidden-layer neural network is said to be locally constant at weight configuration $\theta$ if there exists $\delta>0$ such that
\begin{equation}
\sigma(\mb{w}'^{\top}_i X) = \sigma(\mb{w}^{\top}_i X),\quad \forall \mb{w}_i' \in B(\mb{w}_i, \delta)
\end{equation}
\end{definition}

Definition \ref{def:effective_neruon} and \ref{def:constant_neruon} provide an option to further characterize the neurons when $\Hin_i=0$. ``Locally effective'' is a (strictly) weaker assumption than $\Hin_i\not=0$.
Meanwhile, a locally constant neuron must be a locally ineffective neuron, but not vice versa. To demonstrate the connections among these conditions, we provide the following proposition.

\begin{proposition}\label{prop:condition-connection}
Suppose $(W, \mb{v})$ is a stationary point of the single-hidden-layer neural network. Then the following holds:
    \begin{enumerate}
        \item Suppose the $i$-th neuron of the network satisfies $\Hin_i\not=0$, then it is locally effective.
        \item Suppose the $i$-th neuron of the network is locally constant, then it is not locally effective.
    \end{enumerate}
\end{proposition}

We show that splitting a locally effective neuron always creates saddle points on the embedded plateau, while splitting a neuron that is locally constant, the embedded plateau only contains local minima.

\begin{theorem}
\label{thm:ineffect_creates_saddle}
Suppose that $\ubar{\theta}$ is a local minimum of \narrownet with $\ubar{L}(\ubar{\theta})\not=0$, and $\theta = \neusplit(\ubar{\theta}, \pmb{\lambda})$ is the embedded point of \widenet by neuron splitting. We have the following conclusions.

\begin{enumerate}
    \item If the parent neuron is locally effective at $\ubar{\theta}$, then there exists $\pmb{\lambda} \in \Lambda$ such that $\theta$ is a saddle point of \widenet.
    \item If the parent neuron is locally constant at $\ubar{\theta}$, then $\theta$ is a local minimum of \widenet for any $\pmb{\lambda}\in \Lambda$.
\end{enumerate}
\end{theorem}

\begin{figure}[htbp]
    \centering
    \includegraphics[width=0.8\linewidth]{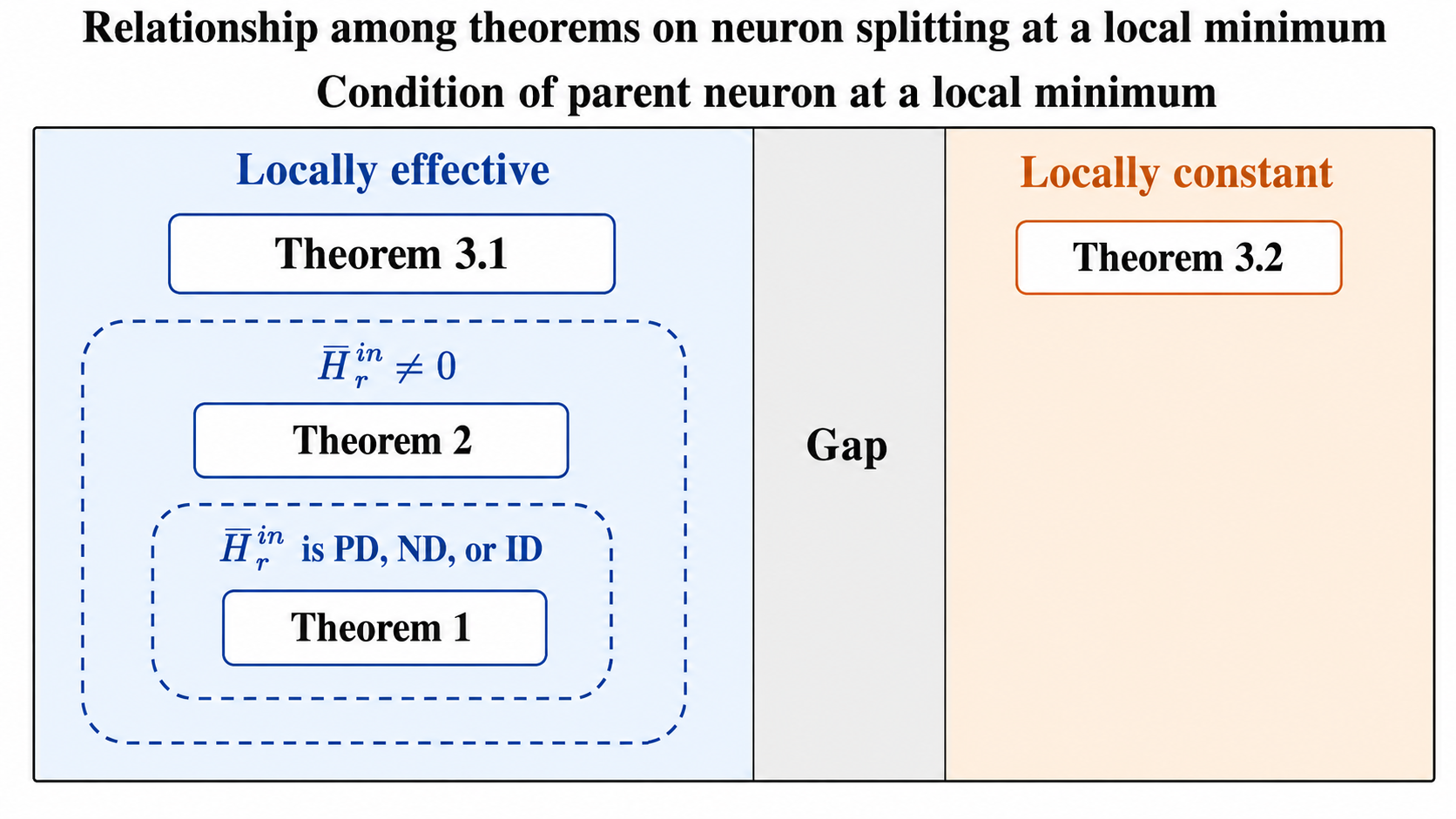}
    \caption{Relationship among theorems on neuron splitting at local minima. The results are categorized based on the condition of the split parent neuron at the local minimum. Specifically, if the parent neuron is locally effective, we obtain stronger results under stronger assumptions.}
    \label{fig:theorem_relation}
\end{figure}

We summarize our results in this section by Table \ref{split-localmin-summary} and Figure \ref{fig:theorem_relation}. We note that for different levels of conditions, we provide  level-matching conclusions. In Theorem \ref{thm:splitting_localmin}, we have the strongest assumptions in $\ubarHin_r$, and we draw the strongest conclusion with a full characterization. As we move downward along the table, the conditions we need for the results become weaker, and consequently, the conclusions we obtain become weaker. Theorem \ref{thm:ineffect_creates_saddle}.1 is the weakest condition we provide so far for a saddle to exist after neuron splitting. We note that there is still a small gap between the conditions of Theorem \ref{thm:ineffect_creates_saddle}.1 and Theorem \ref{thm:ineffect_creates_saddle}.2, which is locally ineffective and locally non-constant.

\begin{table}[htbp]\label{split-localmin-summary}
\centering
\begin{tabular}{lcc}
\toprule
Theorem & Condition & Conclusion \\
\midrule
Theorem \ref{thm:splitting_localmin} & $\ubarHin_r$ is PD, ND or ID & Full characterization \\
Theorem \ref{thm:surely_saddle} & $\ubarHin_r\not=0$ & Positive-measure surely saddle region\\
Theorem \ref{thm:ineffect_creates_saddle}.1 & $\ubarHin_r$ may be $0$. Locally effective & Saddle exists \\
Theorem \ref{thm:ineffect_creates_saddle}.2 & $\ubarHin_r=0$. Locally constant & Saddle does not exist \\
\bottomrule
\end{tabular}
\caption{Summary of the results when splitting neuron at local minimum.}
\end{table}

\subsection{Neuron Splitting at Saddle Point}
In this subsection, we show that by neuron splitting, a saddle point of \narrownet can only give rise to an all-saddle plateau of \widenet.
\begin{theorem}
\label{thm:splitting_saddle}
Let $\ubar{\theta}$ be a saddle point of \narrownet. For any $\pmb{\lambda}\in \Lambda$, $\theta = \neusplit(\ubar{\theta}, \pmb{\lambda})$ is a saddle point of \widenet.
\end{theorem}

\section{Proof Idea: A Perspective of Perturbation Analysis}
\subsection{Weight Perturbation}
\label{sect:weight_perturb}
For a weight configuration $\theta$, let
\begin{equation}
    \theta' = (\mb{v}', W') = (\mb{v}+\Delta \mb{v}, W + \Delta W)
\end{equation}
be a perturbed point around $\theta$. To inspect a stationary point $\theta$ to be a local-min or a saddle, we comparing the values of $L(\theta)$ and $L(\theta')$. Specifically, for an arbitrary $\delta>0$ we define the $\delta$-neighbourhood of $\theta$ as
\begin{equation}
    B(\theta,\delta) = \left\{\theta' = (\mb{v}',W'): \|\mb{v}'-\mb{v}\|^2_2+\|W'-W\|^2_F < \delta^2\right\}.
\end{equation}
If there exists $\delta >0$ such that
\begin{equation}
\label{eq:def_localmin}
L(\theta')-L(\theta) \geq 0 ,~~\forall \theta' \in B(\theta, \delta),      
\end{equation} 
then $\theta$ is a local minimum of $L$. Otherwise, $\theta$ is a saddle point or a local maximum. However, if \eqref{eq:def_localmin} does not hold for any $\delta>0$, $\theta$ is not a locally constant point, hence not a local maximum from Lemma \ref{lem:no_localmax}. Then, $\theta$ must be saddle point. Therefore, 
the problem of characterizing a stationary point boils down to checking whether \eqref{eq:def_localmin} holds for all its neighbourhood.

To this end, we make the following crucial decomposition on the loss difference:
\begin{align}
\label{eq:decompose_empirical_loss1}
    L(\theta')-L(\theta) &= \frac{1}{2}\|\bhat{y}'-\mb{y}\|^2_2 - \frac{1}{2}\|\bhat{y}-\mb{y}\|^2_2 \nonumber\\
    &= \frac{1}{2}\|\bhat{y}'-\bhat{y}\|^2_2 + \langle\bhat{y}-\mb{y}, \bhat{y}'-\bhat{y}\rangle
\end{align}
where $\bhat{y}'=[f(\mb{x}^{(1)};\theta'), \cdots, f(\mb{x}^{(n)};\theta')]^\top \in \realvec{n}$ denotes the network output vector at $\theta'$. Because the network output is the weighted sum of the output of hidden neurons, we have 
\begin{equation}
\label{eq:decompose_empirical_loss2}
    L(\theta')-L(\theta) = \frac{1}{2}\left\|\sum^m_{i=1} \Delta \mb{t}_i\right\|^2_2 + \sum^m_{i'=1} \left\langle \Delta \mb{y}, \Delta \mb{t}_{i'} \right\rangle
\end{equation}
where $\Delta \mb{t}_i^\top = v'_i\sigma(\mb{w}_i'^\top X) - v_i\sigma(\mb{w}_i^\top X) \in \realmat{1}{n}$ denotes the difference of the network output contributed by the $i$-th hidden neuron, and $\Delta \mb{y} \triangleq \bhat{y}-\mb{y}$.

\subsection{Proving local minima: restricting perturbation directions}\label{sec:proofidea_localmin}

To prove that $\theta$ is a local minimum in the wide network, we show that the perturbation directions are restricted in a two-step approach. Specifically, for each perturbed point $\theta'$ in the wide network, we introduce an ``auxiliary pre-split'' perturbed point $\ubar{\theta}'$ in the narrow network. By carefully design the weights of $\ubar{\theta}'$, we can restrict $\ubar{\theta}'$ in the $\delta$-neighborhood of $\ubar{\theta}$. By the assumptions, $\ubar{\theta}$ is a local minimum, so we have
\begin{equation}
\label{eq:def_localmin_narrow}
L(\ubar{\theta}')-L(\ubar{\theta}) \geq 0 ,~~\forall \ubar{\theta}' \in B(\ubar{\theta}, \delta),      
\end{equation}
which implies that the ``pre-split'' perturbation direction is restricted.

The remaining is to prove that the ``post-split'' perturbation direction is restricted. In particular, we aim to prove that the loss difference after splitting is even larger, that is
\begin{equation}
\label{eq:localmin_postsplit}
L(\theta')-L(\theta) \geq L(\ubar{\theta}')-L(\ubar{\theta}) ,~~\forall \theta' \in B(\theta, \delta),      
\end{equation}

Combining \eqref{eq:def_localmin_narrow} and \eqref{eq:localmin_postsplit}, we have 
\begin{equation}
\label{eq:def_localmin_wide}
L(\theta')-L(\theta) \geq 0 ,~~\forall \theta' \in B(\theta, \delta),      
\end{equation}
indicating $\theta$ is a local minimum.

\subsection{Proving saddles: finding decreasing directions}

To prove that $\theta$ is a saddle in the wide network, we find perturbation directions that lead to a decrease of the loss. Unlike in Section \ref{sec:proofidea_localmin} where we need to restrict perturbation directions in both the ``pre-split'' and the ``post-split'' step, this time we simply need to find decreasing directions in either step. Till this end, we prove the results in two cases:

\textbf{Case 1}: If the ``pre-split'' point $\ubar{\theta}$ is a saddle, then we readily have decreasing directions in the ``pre-split'' step, meaning there exists $\ubar{\theta}'$ in an arbitrary small $\delta$-neighborhood of $\ubar{\theta}$ such that $L(\ubar{\theta}')-L(\ubar{\theta})<0$. In this case, we construct the corresponding ``post-split'' perturbed point $\theta'$ based on $\ubar{\theta}'$. Then we have
\begin{equation}
    L(\theta')=L(\ubar{\theta}')<L(\ubar{\theta})=L(\theta).
\end{equation}
This means that there exists $\theta'$ in an arbitrary small $\delta$-neighborhood of $\theta$ such that $L(\theta')-L(\theta) < 0$, implying that $\theta$ is a saddle.

\textbf{Case 2}: If the ``pre-split'' point $\ubar{\theta}$ is a local-min, we aim to find decreasing directions in the ``post-split'' step. To find the decreasing direction, note that by the decomposition \eqref{eq:decompose_empirical_loss2}, the loss difference is decomposed into two parts:
\begin{equation}
\label{eq:decompose_empirical_loss3}
    L(\theta')-L(\theta) = \underbrace{\frac{1}{2}\left\|\sum^m_{i=1} \Delta \mb{t}_i\right\|^2_2}_{\text{I}} + \underbrace{\sum^m_{i'=1} \left\langle \Delta \mb{y}, \Delta \mb{t}_{i'} \right\rangle}_{\text{II}}
\end{equation}

Given appropriate conditions on the inner Hessian, we can construct $\theta'$ in an arbitrary small $\beta$-neighborhood of $\theta$ such that
\begin{equation}
    \label{eq:decompose_magnitude}
    \text{II}<0, ~~ |\text{II}| = O(\beta^2), ~~|\text{I}| = o(\beta^2).
\end{equation}
Combining \eqref{eq:decompose_empirical_loss3} and \eqref{eq:decompose_magnitude}, we have $L(\theta')-L(\theta)=\text{I}+\text{II}<0$, implying that $\theta$ is a saddle.

\section{Proof of Theorems}
\subsection{Proof of Theorem \ref{thm:splitting_localmin}}
\label{sect:pfThmSL}
The proof of Theorem \ref{thm:splitting_localmin} relies on a quadratic approximation of neuron output. For an perturbed weight configuration $\theta'$, we considered the perturbed output $\sigma(\mb{w}_i'^\top X)$ of the $i$-th neuron. We perform Taylor expansion with respect to $\mb{w}_i$ as
\begin{multline}
    \sigma(\mb{w}_i'^\top X) = \mb{z}_i^\top + \Delta \mb{w}_i^\top  \left(\mb{z}_i'^\top\circ X\right) \\ + \frac{1}{2}\sum^{d}_{j=1}\sum^{d}_{j'=1}\Delta w_{i,j} \Delta w_{i,j'} (\mb{z}_i''\circ \mb{x}_j \circ \mb{x}_{j'})^\top
    + \mb{o}^\top(\|\Delta \mb{w}_i\|^2_2)
\end{multline}
where with some abuse of notation we denote
\begin{equation}
\label{eq:pfThmSL_def_vec_had_matrix}
    \mb{z}_i'^\top\circ X = \begin{bmatrix}
(\mb{z}_i'\circ \mb{x}_1)^\top \\
(\mb{z}_i'\circ \mb{x}_2)^\top \\
\vdots\\
(\mb{z}_i'\circ \mb{x}_d)^\top
\end{bmatrix}
\in \realmat{d}{n},
\end{equation}
and $\mb{o}(\cdot)$ denotes an $n$-dimensional infinitesimal vector with $\lim_{t\rightarrow 0}\|\mb{o}(t)\|_2 = 0$. This allows us to rewrite $\Delta \mb{t}_i$ as
\begin{align}
\label{eq:pfThmSL_decompose_delta_t}
    \Delta \mb{t}_i^\top =~& v_i' \sigma(\mb{w}_i'^\top X) - v_i  \mb{z}_i'^\top \nonumber\\
    =~ &(v_i + \Delta v_i ) \Delta \mb{w}_i^\top  \left(\mb{z}_i'^\top\circ X\right) + \frac{1}{2}(v_i + \Delta v_i ) \sum^{d}_{j=1}\sum^{d}_{j'=1}\Delta w_{i,j} \Delta w_{i,j'} (\mb{z}_i''\circ \mb{x}_j \circ \mb{x}_{j'})^\top \nonumber\\
    &+ (v_i + \Delta v_i ) \mb{o}^\top(\|\Delta \mb{w}_i\|^2_2) + \Delta v_i \mb{z}_i^\top.
\end{align}

If $\theta$ is a stationary point of the empirical loss $L$, we have
\begin{subequations}
\label{eq:pfThmSL_stationary_cond}
\begin{gather}
    \frac{\partial L}{\partial v_i} = \langle \Delta \mb{y}, \mb{z}_i\rangle = 0 \\
    \label{eq:pfThmSL_stationary_cond_w}
    \frac{\partial L}{\partial \mb{w}_i} =
    v_i \begin{bmatrix}
    \langle \Delta \mb{y}, \mb{z}_i' \circ \mb{x}_1 \rangle \\
    \langle \Delta \mb{y}, \mb{z}_i' \circ \mb{x}_2 \rangle \\
    \vdots \\
    \langle \Delta \mb{y}, \mb{z}_i' \circ \mb{x}_d \rangle 
    \end{bmatrix}
    = \mb{0}.   
\end{gather}
\end{subequations}
If $v_i \not=0$, \eqref{eq:pfThmSL_stationary_cond_w} implies $\langle \Delta \mb{y}, \mb{z}_i' \circ \mb{x}_j \rangle = 0$ for all $j\in [d]$. Then, if $v_i \not= 0$, combining  \eqref{eq:pfThmSL_decompose_delta_t} and \eqref{eq:pfThmSL_stationary_cond} we obtain
\begin{align}
\label{eq:pfThmSL_decompose_delta_t2}
    & ~\langle \Delta \mb{y}, \Delta \mb{t}_i \rangle \nonumber \\
    =& ~ \frac{1}{2}  (v_i + \Delta v_i )\sum^{d}_{j=1}\sum^{d}_{j'=1}\Delta w_{i,j} \Delta w_{i,j'} \langle \Delta \mb{y} , \mb{z}_i''\circ \mb{x}_j \circ \mb{x}_{j'} \rangle + (v_i+\Delta v_i) \cdot \langle \Delta \mb{y}, \mb{o}(\|\Delta \mb{w}_i\|^2_2)\rangle \nonumber \\
    =& ~ \frac{1}{2}\left(1+\frac{\Delta v_i}{v_i}\right)\Delta \mb{w}_i^\top \Hin_i \Delta\mb{w}_i + (v_i+\Delta v_i)\cdot \langle \Delta \mb{y}, \mb{o}(\|\Delta \mb{w}_i\|^2_2)\rangle
\end{align}
where $\Hin_i$ is the inner Hessian matrix of the $i$-th neuron defined by \eqref{eq:def_inner_Hessian}.

The remaining proof is divided into two parts. In the first part we prove that the embedded point is a local-min for any $\lambda\in \Lambda_{\mathrm{PD}}$ ($\lambda\in \Lambda_{\mathrm{ND}}$) if the parent neuron has a PD (ND) inner Hessian. In the second part we prove the embedded point to be a saddle in the remaining region of $\Lambda$.

\subsubsection{Proof of Theorem \ref{thm:splitting_localmin}--- Part I: Proving Local-Min for PD inner Hessian}
\label{sect:pfThmSL_PD_localmin}
We consider the case of PD inner Hessian and $\pmb{\lambda}\in \Lam{PD}$. By definition of the inner Hessian matrix in \eqref{eq:def_inner_Hessian},
$\lambda_{\min}(\ubarHin_r)>0$ implies $\ubar{v}_r \not= 0$. Further, for $\pmb{\lambda} \in \Lambda_{\mathrm{PD}}$ we have $\lambda_j \in (0,1)$, $\forall j\in[0:(m-r)]$. This implies that $v_{r+j} = \lambda_j \ubar{v}_r \not=0$, and hence $v'_{r+j}\not=0$ for any perturbed point $\theta' \in B(\theta,\delta)$ with sufficiently small $\delta$. Define ``perturbed'' versions of $\{\lambda_j\}^{m-r}_{j=0}$ as
\begin{equation}
    \label{eq:pfThmSL_PD_localmin_def_lamprime}
    \lambda'_j = \frac{v_{r+j}'}{\sum^{m-r}_{k=0}v_{r+k}'}, ~~ j\in [0:(m-r)].
\end{equation}
We readily have $\sum^{m-r}_{j=0}\lambda'_j = 1$. If $\delta$ is sufficiently small, $\lambda'_j$ is close to $\lambda_j$, and we have $\lambda_j \in (0,1)$. In the remaining proof of Section \ref{sect:pfThmSL_PD_localmin}, we always assume $\lambda_j' \in (0,1)$ for all $j\in [0:(m-r)]$.

Now, for any $\theta' \in B(\theta,\delta)$ we define an ``auxiliary'' perturbed point $\ubar{\theta}' = (\ubar{\mb{v}}',\ubar{W}') = (\ubar{\mb{v}}+\Delta \ubar{\mb{v}},\ubar{W}+\Delta \ubar{W})$ for \narrownet, given by
\begin{subequations}
\label{eq:pfThmSL_PD_localmin_auxiliary_perturb}
\begin{gather}
    \Delta\ubar{\mb{w}}_i = \Delta \mb{w}_i,~~\Delta \ubar{v}_i = \Delta v_i,\quad i\in[r-1]\\
    \Delta\ubar{\mb{w}}_r = \sum^{m}_{i'=r}\lambda'_{i'-r}\Delta \mb{w}_{i'},\quad \Delta \ubar{v}_r =\sum^{m}_{i'=r} \Delta v_{i'}.
\end{gather}
\end{subequations}
That is, the perturbation of the hidden weight at the parent neuron is the weighted average of the hidden weights at the daughter neurons, and the perturbation of the output weight at the parent neuron of is the sum of the output weights at the daughter neurons. Notice that $\lambda_j'\in (0,1)$ for all $j\in[0:(m-r)]$. By Cauchy–Schwarz inequality we have
\begin{subequations}
\label{eq:pfThmSL_PD_localmin_perturb_range}
    \begin{gather}
        \|\Delta \ubar{\mb{v}}\|^2_2 = \sum^{r}_{i=1} \Delta \ubar{v}_i^2
        \leq
        \sum^{r-1}_{i=1} \Delta v_i^2 + (m-r+1)\sum^m_{i'=r}  \Delta v_{i'}^2  \leq  
        (m-r+1)\|\Delta \mb{v}\|_2^2 \\
        \|\Delta \ubar{W}\|_F^2 = 
        \sum^{r}_{i=1} \|\Delta\ubar{ \mb{w}}_i\|_2^2 \leq \sum^{r-1}_{i=1} \|\Delta \mb{w}_i\|_2^2 + \sum^{m-r}_{j=0} |\lambda_j'|^2 \sum^m_{i'=r} \|\Delta \mb{w}_{i'}\|_2^2 \leq \|\Delta W\|_F^2.
    \end{gather}
\end{subequations}
Denoting $C_1=\sqrt{m-r+1}$, we have $\ubar{\theta}'\in B(\ubar{\theta},C_1\delta)$ as long as $\theta' \in B(\theta,\delta)$.

For \narrownet and the auxiliary perturbation $\ubar{\theta}'$, denote the change of network output contributed by the $i$-th neuron by
\begin{equation}
\label{eq:pfThmSL_PD_localmin_delta_ubart}
    \Delta \ubar{\mb{t}}_i^\top = \ubar{v}_i'\sigma(\ubar{\mb{w}}'^{\top}_i X)- \ubar{v}_i \ubar{\mb{z}}_i^\top \in \realmat{1}{n}, \quad i \in [r].
\end{equation}
From \eqref{eq:pfThmSL_PD_localmin_auxiliary_perturb} we have $\Delta \mb{t}_i = \Delta \ubar{\mb{t}}_i$ for any $i\in[r-1]$, yielding
\begin{align}
\label{eq:pfThmSL_PD_localmin_output_difference}
\bhat{y}' - \bhat{y} =&~ \sum^m_{i=1}\Delta \mb{t}_i = \sum^{r-1}_{i=1}  \Delta \ubar{\mb{t}}_i +
\sum^m_{i'=r}\Delta \mb{t}_{i'}\nonumber \\
= &~ \sum^r_{i=1} \Delta \ubar{\mb{t}}_i + \sum^m_{i'=r}\Delta \mb{t}_{i'} - \Delta \ubar{\mb{t}}_r\nonumber \\
= &~ \left(\ubar{\bhat{y}}' - \ubar{\bhat{y}}\right) + \mb{e}
\end{align}
where $\mb{e}\in \realvec{n}$ is defined to be
\begin{align}
\label{eq:pfThmSL_PD_localmin_def_e}
    \mb{e}^\top &=  \sum^m_{i=r}\Delta \mb{t}_i^\top - \Delta\ubar{\mb{t}}_r^\top \nonumber \\
    & = \sum_{i=r}^m \left[v_i'\sigma(\mb{w}_i'^\top X)-v_i\sigma(\mb{w}_i^\top X)\right] - \left[ \ubar{v}_r'\sigma(\ubar{\mb{w}}_r'^\top X) - \ubar{v}_r\sigma(\ubar{\mb{w}}_r^\top X)\right] \nonumber \\
    &= \left[\sum_{i=r}^m v_i'\sigma(\mb{w}_i'^\top X) - \ubar{v}_r'\sigma(\ubar{\mb{w}}_r'^\top X)\right] - \left[\sum_{i'=r}^m v_{i'}\sigma(\mb{w}_{i'}^\top X) - \ubar{v}_r\sigma(\ubar{\mb{w}}_r^\top X)\right] \nonumber \\
    &  = \sum_{i=r}^m v_i'\sigma(\mb{w}_i'^\top X) - \ubar{v}_r'\sigma(\ubar{\mb{w}}_r'^\top X).
\end{align}
where the last equality follows from the fact that $\ubar{v}_r = \sum^m_{i'=r} v_{i'}$ and $\mb{w}_i' = \ubar{\mb{w}}_r$, $\forall i' \in [r:m]$. Then, by decomposition \eqref{eq:decompose_empirical_loss1} we have
\begin{align}
\label{eq:pfThmSL_PD_localmin_loss_difference}
    L(\theta')-L(\theta) &= \frac{1}{2}\left\|\bhat{y}'-\bhat{y}\right\|^2_2 + \langle\Delta \mb{y}, \bhat{y}'-\bhat{y}\rangle \nonumber\\
    &= \frac{1}{2}\left\|\ubar{\bhat{y}}'-\ubar{\bhat{y}} + \mb{e}\right\|^2_2 +  \left\langle\Delta \ubar{\mb{y}}, \ubar{\bhat{y}}'-\ubar{\bhat{y}}\right\rangle + \langle \Delta \mb{y}, \mb{e}\rangle.
\end{align}
We see that the difference of empirical loss is relevant to the term $\mb{e}$. Define
\begin{equation}
\label{eq:pfThmSL_PD_localmin_def_tilw}
    \Delta\btil{w}_i \triangleq \Delta \mb{w}_i - \Delta \ubar{\mb{w}}_r, ~~i\in [r:m].
\end{equation}
We provide the following lemma for bounding the terms related to $\mb{e}$, whose proof is deferred to \ref{app:pfLemPDL}.
\begin{lemma}
\label{lem:PD_localmin}
Suppose that $\lambda_{\min}(\ubarHin_r)>0$ and $\pmb{\lambda}\in \Lam{PD}$. For any $\epsilon>0$, there exists $\delta>0$ such that
\begin{subequations}
\label{eq:lemPDL_ineq}
\begin{gather}
\label{eq:lemPDL_ineq1}
    \left\|(\ubar{\bhat{y}}'-\ubar{\bhat{y}}) + \mb{e}\right\|^2_2 \geq\left\|\ubar{\bhat{y}}'-\ubar{\bhat{y}}\right\|^2_2  -\epsilon \sum^{m-r}_{j=0}\lambda_{j}\|\Delta \btil{w}_{r+j}\|^2_2 \\
\label{eq:lemPDL_ineq2}
    \langle \Delta \mb{y}, \mb{e} \rangle \geq  \frac{1}{4}\lambda_{\min}(\ubarHin_r)\sum^{m-r}_{j=0}\lambda_{j}\|\Delta \btil{w}_{r+j}\|^2_2 \geq 0
\end{gather}
\end{subequations}
for all $\theta' \in B(\theta,\delta)$.
\end{lemma}

Setting $\epsilon =\lambda_{\min}(\ubarHin_r)/4$ in Lemma \ref{lem:PD_localmin}, there exists $\delta>0$ such that for any $\theta'\in B(\theta,\delta)$ we can bound the difference of empirical loss as
\begin{align}
    L(\theta')-L(\theta) &= \frac{1}{2}\left\|(\ubar{\bhat{y}}'-\ubar{\bhat{y}}) + \mb{e}\right\|^2_2 +  \left\langle\Delta \ubar{\mb{y}}, \ubar{\bhat{y}}'-\ubar{\bhat{y}}\right\rangle
    + \langle \Delta \mb{y}, \mb{e}\rangle \nonumber \\
    & \geq \frac{1}{2}\left\|\ubar{\bhat{y}}'-\ubar{\bhat{y}}\right\|^2_2 +\left\langle\Delta \ubar{\mb{y}}, \ubar{\bhat{y}}'-\ubar{\bhat{y}}\right\rangle + \frac{1}{8}\lambda_{\min}(\ubarHin_r)\sum^{m-r}_{j=0}\lambda_{j}\|\Delta \btil{w}_{r+j}\|^2_2 \nonumber \\
    &\geq\frac{1}{2}\left\|\ubar{\bhat{y}}'-\ubar{\bhat{y}}\right\|^2_2 +\left\langle\Delta \ubar{\mb{y}}, \ubar{\bhat{y}}'-\ubar{\bhat{y}}\right\rangle \nonumber \\
    & = \ubar{L}(\theta')-\ubar{L}(\theta)
\end{align}
where the last equality follows from decomposition \ref{eq:decompose_empirical_loss1} for \narrownet. Because $\ubar{\theta}$ is a local minimum of \narrownet, we have $\ubar{L}(\ubar{\theta}')-\ubar{L}(\ubar{\theta}) \geq 0 $ for sufficiently small $\delta$, which implies $L(\theta')-L(\theta)\geq 0$ for \widenet. Thus, $\theta = \neusplit(\ubar{\theta}, \pmb{\lambda})$ is a local minimum of \widenet for $\pmb{\lambda}\in\Lam{PD}$.

\subsubsection{Proof of Theorem \ref{thm:splitting_localmin}--- Part II: Proving Local-Min for ND inner Hessian}
\label{sect:pfThmSL_ND_localmin}
We consider the case of ND inner Hessian and $\lambda\in \Lam{ND}$. Without loss of generality, we assume $\lambda_0>0$ and $\lambda_{j}<0$ for all $j\in[1:(m-r)]$. This proof consists of two steps. First, we prove the case with $m = r+1$, i.e., splitting the parent neuron into two daughter neurons only. Then, we extend the proof to the case with $m>r+1$.
   
\paragraph{Step 1: Proof for $m=r+1$.} In this case, we have $\lambda_0>1$ and $\lambda_1<0$ for $\pmb{\lambda}\in \Lam{ND}$. The proof of Step 1 follows a similar idea with Section~\ref{sect:pfThmSL_PD_localmin}, but we only need to handle a single-step  splitting here.

By the definition of inner Hessian in \eqref{eq:def_inner_Hessian},
$\lambda_{\max}(\ubarHin_r)<0$ implies $\ubar{v}_r\not= 0$. Further, because $\lambda_j\not=0$ for $\pmb{\lambda} \in \Lam{ND}$, we have $v_{r+j} = \lambda_j \ubar{v}_r \not= 0$. For a perturbed point $\theta' \in B(\theta,\delta)$ of \widenet, we define
\begin{equation}
    \label{eq:pfThmSL_ND_localmin_lamprime}
    \lambda_{0}' = \frac{v_{r}'}{v_{r}'+v_{r+1}'}, ~~ \lambda_{1}' = \frac{v_{r+1}'}{v_{r}'+v_{r+1}'}.
\end{equation}
We readily have $\lambda_0' + \lambda_1' = 1$. Further, $\lambda'_j$ converges to $\lambda_j$ as $\delta \rightarrow 0$, $j\in\{0,1\}$. In the remaining proof of Step 1, we always assume a sufficiently small $\delta$ such that
\begin{equation}
\label{eq:pfThmSL_ND_localmin_lambdaprime_condition}
    0<\frac{1}{2}\lambda_0 < \lambda_0' < 2\lambda_0, \quad  2\lambda_1 < \lambda_1'< \frac{\lambda_1}{2}< 0. 
\end{equation}

Now, for any $\theta' \in B(\theta,\delta)$ we define an ``auxiliary'' perturbed point $\ubar{\theta}' = (\ubar{\mb{v}}',\ubar{W}') = (\ubar{\mb{v}}+\Delta \ubar{\mb{v}},\ubar{W}+\Delta \ubar{W})$ for \narrownet, given by
\begin{subequations}
\label{eq:pfThmSL_ND_localmin_auxiliary_perturb}
\begin{gather}
    \Delta\ubar{\mb{w}}_i = \Delta \mb{w}_i,~~\Delta \ubar{v}_i = \Delta v_i,~~i\in[r-1]\\
    \Delta\ubar{\mb{w}}_r = \lambda_{0}'\Delta \mb{w}_r + \lambda_{1}'\Delta \mb{w}_{r+1},~~\Delta \ubar{v}_r = \Delta  v_{r}+ \Delta v_{r+1}.
\end{gather}
\end{subequations}
Note that $|\lambda_0'|= |1-\lambda_1'|=1+|\lambda_1'|>\max\{|\lambda_1'|,1\}$. By Cauchy–Schwarz inequality
\begin{subequations}
\label{eq:pfThmSL_ND_localmin_perturbation_range}
    \begin{gather}
        \|\Delta \ubar{\mb{v}}\|^2_2 = \sum^{r}_{i=1} \Delta \ubar{v}_i^2
        \leq
        \sum^{r-1}_{i=1} \Delta v_i^2 + 2\sum^{r+1}_{i'=r}  \Delta v_{i'}^2  \leq
        2\|\Delta \mb{v}\|_2^2 \\
        \label{eq:pfThmSL_ND_localmin_perturbation_range2}
        \|\Delta \ubar{W}\|_F^2 = 
        \sum^{r}_{i=1} \|\Delta\ubar{ \mb{w}}_i\|_2^2 \leq \sum^{r-1}_{i=1} \|\Delta \mb{w}_i\|_2^2 + \sum^{1}_{j=0} \lambda_j'^2\sum^{r+1}_{i'=r} \|\Delta \mb{w}_{i'}\|_2^2 \leq 8\lambda_0^2\|\Delta W\|_F^2.
    \end{gather}
\end{subequations}
where the last inequality of \eqref{eq:pfThmSL_ND_localmin_perturbation_range2} follows from $ |\lambda_1'|< \lambda_0' < 2\lambda_0$. Denoting $C_2 = 2\sqrt{2}\lambda_0$, we have $\ubar{\theta}'\in B(\ubar{\theta},C_2\delta)$ as long as $\theta' \in B(\theta,\delta)$.

Following a similar analysis from \eqref{eq:pfThmSL_PD_localmin_delta_ubart}-\eqref{eq:pfThmSL_PD_localmin_loss_difference}, we have
\begin{align}
    L(\theta')-L(\theta) = \frac{1}{2}\left\|\ubar{\bhat{y}}'-\ubar{\bhat{y}} + \mb{e}\right\|^2_2 +  \left\langle\Delta \ubar{\mb{y}}, \ubar{\bhat{y}}'-\ubar{\bhat{y}}\right\rangle + \langle \Delta \mb{y}, \mb{e}\rangle.
\end{align}
where $\mb{e}\in \realvec{n}$ is defined to be
\begin{equation}
\label{eq:pfThmSL_ND_localmin_def_e}
    \mb{e}^\top   = v_r'\sigma(\mb{w}_r'^\top X)  + v_{r+1}'\sigma(\mb{w}_{r+1}'^\top X) - \ubar{v}_r'\sigma(\ubar{\mb{w}}_r'^\top X).
\end{equation}
Subsequently, we define
\begin{equation}
\label{eq:pfThmSL_ND_localmin_def_ub}
    \beta_0 \triangleq \frac{1}{\lambda_0}, \quad \beta_1 \triangleq \frac{|\lambda_1|}{\lambda_0}, \quad \mb{u}_0 \triangleq \Delta \ubar{\mb{w}}_r - \Delta \mb{w}_r, \quad \mb{u}_1 \triangleq \Delta \mb{w}_{r+1} - \Delta \mb{w}_r
\end{equation}
and provide the following lemma whose proof is deferred to Appendix \ref{app:pfLemNDL}.
\begin{lemma}
\label{lem:ND_localmin}
Suppose that $m=r+1$, $\lambda_{\max}(\ubarHin_r)<0$, and $\pmb{\lambda}\in \Lam{ND}$. For any $\epsilon>0$, there exists $\delta>0$ such that
\begin{subequations}
\begin{gather}
\label{eq:lemma_ND_localmin_ineq1}
    \left\|(\ubar{\bhat{y}}'-\ubar{\bhat{y}}) + \mb{e}\right\|^2_2 \geq\left\|\ubar{\bhat{y}}'-\ubar{\bhat{y}}\right\|^2_2  -\epsilon \left(\beta_0 \|\mb{u}_0\|^2_2 + \beta_1 \|\mb{u}_1\|^2_2\right)  \\
\label{eq:lemma_ND_localmin_ineq2}
    \langle \Delta \mb{y}, \mb{e} \rangle \geq  \frac{1}{8}\left|\lambda_{\max}(\ubarHin_r)\right|\cdot \left(\beta_0 \|\mb{u}_0\|^2_2 + \beta_1 \|\mb{u}_1\|^2_2\right) \geq 0
\end{gather}
\end{subequations}
for all $\theta' \in B(\theta,\delta)$.
\end{lemma}

Setting $\epsilon = |\lambda_{\max}(\ubarHin_r)|/8$ in Lemma \ref{lem:ND_localmin}, there exists $\delta>0$ such that for any $\theta'\in B(\theta,\delta)$ we can bound the difference of empirical loss as
\begin{align}
    L(\theta')-L(\theta) &= \frac{1}{2}\left\|(\ubar{\bhat{y}}'-\ubar{\bhat{y}}) - \mb{e}\right\|^2_2 +  \left\langle\Delta \ubar{\mb{y}}, \ubar{\bhat{y}}'-\ubar{\bhat{y}}\right\rangle
    + \langle \Delta \mb{y}, \mb{e}\rangle \nonumber \\
    & \geq \frac{1}{2}\left\|\ubar{\bhat{y}}'-\ubar{\bhat{y}}\right\|^2_2 +\left\langle\Delta \ubar{\mb{y}}, \ubar{\bhat{y}}'-\ubar{\bhat{y}}\right\rangle + \frac{1}{16}|\lambda_{\max}(\ubarHin_r)|\cdot \left(\beta_0 \|\mb{u}_0\|^2_2 + \beta_1 \|\mb{u}_1\|^2_2\right) \nonumber \\
    &\geq\frac{1}{2}\left\|\ubar{\bhat{y}}'-\ubar{\bhat{y}}\right\|^2_2 +\left\langle\Delta \ubar{\mb{y}}, \ubar{\bhat{y}}'-\ubar{\bhat{y}}\right\rangle \nonumber \\
    & = \ubar{L}(\ubar{\theta}')-\ubar{L}(\ubar{\theta}).
\end{align}
Since $\ubar{\theta}$ is a local minimum of \narrownet, we have $\ubar{L}(\ubar{\theta}')-\ubar{L}(\ubar{\theta})\geq 0$ for sufficiently small $\delta$, and hence $L(\theta')-L(\theta)\geq 0$. Thus, $\theta$ is a local minimum of \widenet.

\paragraph{Step 2: Proof for $m>r+1$.} In this case, the weight configuration $\theta$ of \widenet can be seen as the result of a two-step neuron-splitting. 

First, we construct a $(r+1)$-neuron network, denoted by \emph{\mediumnet}, by splitting the $r$-th neuron of \narrownet at $\ubar{\theta}$ with coefficients $\pmb{\lambda}\upbrac{1} = [\lambda_0, \lambda_{1:(m-r)}]^\top$, where $\lambda_{1:(m-r)} \triangleq \sum^{m-r}_{j=1}\lambda_j$. The resulting weight configuration of \mediumnet is denoted by $\theta^{\mathrm{med}}$. By the fact $\pmb{\lambda}\in \Lam{ND}$ and the assumption at the beginning of this section, we have $\lambda_0 + \lambda_{1:(m-r)} = 1$ and $\lambda_0>0$, $\lambda_{1:(m-r)}<0$. This meets the condition of Step 1. We directly apply the conclusion of Step 1 and obtain that $\theta^{\mathrm{medium}}$ is a local minimum of \mediumnet.

Next, we note that $\theta = \neusplit(\theta, \pmb{\lambda})$ can be seen as an embedded point by splitting the $(r+1)$-neuron of \mediumnet at $\theta^{\mathrm{medium}}$ with coefficients
\begin{equation}
    \pmb{\lambda}\upbrac{2} = \left[\lambda\upbrac{2}_1, \lambda\upbrac{2}_2, \cdots, \lambda\upbrac{2}_{m-r}\right]^\top =   \left[\frac{\lambda_1}{\lambda_{1:(m-r)}}, \frac{\lambda_2}{\lambda_{1:(m-r)}}, \cdots, \frac{\lambda_{m-r}}{\lambda_{1:(m-r)}}\right]^\top.
\end{equation}
Clearly $\sum^{m-r}_{j=1} \lambda\upbrac{2}_j = 1$. Noting that we assume $\lambda_{j}<0$ for all $j\in[1:(m-r)]$ at the beginning of this section, we have $\lambda\upbrac{2}_j > 0$ for all $j\in[1:(m-r)]$. Further, let $H^{\mathrm{in,med}}_{r+1}$ be the inner Hessian matrix (defined similarly as \eqref{eq:def_inner_Hessian}) for the $(r+1)$-th neuron of \mediumnet, we have
\begin{equation}
H^{\mathrm{in,med}}_{r+1} = \lambda_{1:(m-r)} \ubarHin_r. 
\end{equation}
Because $ \ubarHin_r$ is negative definite and $\lambda_{1:(m-r)}<0$, $H^{\mathrm{in,med}}_{r+1}$ is positive definite. Further, notice that $\theta^{\mathrm{med}}$ is a local minimum of \mediumnet. The neuron splitting from \mediumnet to \widenet meets the condition of the second conclusion of Theorem~\ref{thm:splitting_localmin}. We apply the proof in Section~\ref{sect:pfThmSL_PD_localmin} and obtain that $\theta$ is a local minimum of \widenet.

We conclude that if $\ubarHin_r$ is negative definite, $\theta$ is a local minimum of \widenet for any $\pmb{\lambda} \in \Lam{ND}$.

\subsubsection{Proof of Theorem \ref{thm:splitting_localmin}--- Part III: Proving Saddles}
The following lemma provides sufficient conditions for an embedded stationary point to be a saddle. Specifically, we show that as long as the empirical loss is non-zero and the inner Hessian of the parent neuron has non-zero eigenvalues, we can always find coefficients $\pmb{\lambda}$ such that the embedded point is a saddle point. 
\begin{lemma}
\label{lem:saddle_region}
Let $\ubar{\theta}$ be a stationary point of \narrownet with $\ubar{L}(\ubar{\theta})\not=0$. Denote
\begin{subequations}
\label{eq:def_lamda_plus_minus}
\begin{gather}
    \Lam{+}\triangleq \{\pmb{\lambda}\in\Lambda: \exists j\in[0:m-r] \text{ s.t. } \lambda_j\leq 0\}\\
    \Lam{-}\triangleq \{\pmb{\lambda}\in\Lambda: \exists j,j'\in[0:m-r] \text{ s.t. } \lambda_j,\lambda_{j'}\geq 0, ~j\not=j'\}
\end{gather}
\end{subequations}
We have that
\begin{enumerate}
    \item If $\lambda_{\max}(\ubarHin_r)>0$, $\theta = \neusplit(\ubar{\theta}, \pmb{\lambda})$ is a saddle point of \widenet for all $\lambda\in \Lam{+}$
    \item If $\lambda_{\min}(\ubarHin_r)<0$, $\theta = \neusplit(\ubar{\theta}, \pmb{\lambda})$ is a saddle point of \widenet for all $\lambda\in \Lam{-}$.
\end{enumerate}
\end{lemma}
The key idea of proving Lemma \ref{lem:saddle_region} is to perturb the embedded point along the direction associated with a non-zero eigenvalue of the inner Hessian, and then show that empirical loss will decrease. For positive and negative eigenvalues of the inner Hessian, we construct two different perturbations, and thus we identify two different regions on the embedded plateau in Lemma \ref{lem:saddle_region}. We defer the proof of Lemma \ref{lem:saddle_region} to Appendix \ref{app:pfLemSR}.

If the inner Hessian of the parent neuron is positive definite, we have $\lambda_{\max}(\ubarHin_r)>0$. We apply the first conclusion of Lemma \ref{lem:saddle_region}. Notice that $\Lam{+} = \Lambda \backslash\Lam{PD}$. We conclude that $\theta$ is a saddle point of \widenet for all $\pmb{\lambda}\in \Lambda \backslash \Lam{PD}$.

If the inner Hessian of the parent neuron is positive definite, we have $\lambda_{\min}(\ubarHin_r)<0$. We apply the second conclusion of Lemma \ref{lem:saddle_region}. Notice that $\Lam{-} = \Lambda \backslash\Lam{ND}$. We conclude that $\theta$ is a saddle point of \widenet for all $\pmb{\lambda}\in \Lambda \backslash \Lam{ND}$.

If the inner Hessian of the parent neuron is indefinite, we have both $\lambda_{\max}(\ubarHin_r)>0$ and $\lambda_{\min}(\ubarHin_r)<0$. We apply the both conclusions of Lemma \ref{lem:saddle_region}. Notice that for any $m>r$, $\Lambda = \Lam{+}\cup \Lam{-}$. We conclude that $\theta$ is a saddle point of \widenet for all $\pmb{\lambda}\in \Lambda$.

Combined with Section~\ref{sect:pfThmSL_PD_localmin} and \ref{sect:pfThmSL_ND_localmin}, we complete the proof of Theorem \ref{thm:splitting_localmin}.

\subsection{Proof of Theorem \ref{thm:surely_saddle}}
Because $\ubarHin_r\not=\mb{0}$, at least one of $\lambda_{\max} \ubarHin_r >0$ and $\lambda_{\min} \ubarHin_r <0$ holds. By Lemma \ref{lem:saddle_region}, $\theta$ is a saddle point of \widenet for any $\pmb{\lambda} \in \Lam{+}\cap\Lam{-}$. To prove Theorem \ref{thm:surely_saddle}, it suffices to show that $\Lam{+}\cap\Lam{-} = \Lambda\backslash \Lam{PD}\cup\Lam{ND}$. By the definitions in \eqref{eq:def_lambda_pd_nd} and \eqref{eq:def_lamda_plus_minus}, we directly have $\Lam{PD} = \Lambda \backslash \Lam{+}$ and $\Lam{ND} = \Lambda \backslash \Lam{-}$. Thus, $\Lam{+}\cap\Lam{-} = \Lambda\backslash \Lam{PD}\cup\Lam{ND}$. We complete the proof.

\subsection{Proof of Theorem \ref{thm:ineffect_creates_saddle}}
\subsubsection{Neuron Splitting at Locally Effective Neuron}
Assume that the parent neuron is locally effective. Consider an arbitrary $\delta>0$. By Definition \ref{def:effective_neruon}, there exists $\mb{u}_0\in \realvec{d}$ with $\|\mb{u}_0\|_2<\delta$ such that
\begin{equation}
\label{eq:pfThmICS_nonzero}
    \left\langle \Delta \mb{y},\sigma^\top\left((\ubar{\mb{w}}_r+\mb{u}_0)^\top X\right) \right\rangle \not = 0
\end{equation}
where $\Delta \mb{y} \triangleq \bhat{y}-\mb{y} = \ubar{\bhat{y}}-\mb{y}$.

Consider splitting coefficients $\pmb{\lambda}\in \Lambda$ with $\lambda_0 = 0$, and let $\theta = \neusplit(\ubar{\theta}, \pmb{\lambda})$ be the embedded point of \widenet. We construct a perturbed point $\theta' = (\mb{v}', W')$ as
\begin{subequations}
\label{eq:pfThmICS_pert}
\begin{gather}
    v'_r = v_r + \alpha, \quad \mb{w}'_r = \mb{w}_r + \mb{u}_0 \\
    v'_i = v_i, \quad \mb{w}'_i = \mb{w}_i, \quad i \in [m]\backslash\{r\}
\end{gather}
\end{subequations}
where $\alpha$ is a scalar whose value will be determined later. Clearly, $\theta' \in B(\theta, \delta)$ if
\begin{equation}
0<|\alpha|<\sqrt{\delta^2 - \|\mb{u}_0\|_2^2}.
\end{equation}

Let $\Delta \mb{t}_i^\top = v'_i\sigma(\mb{w}_i'^\top X) - v_i\sigma(\mb{w}_i^\top X) \in \realmat{1}{n}$ denote the difference of network output contributed by the $i$-th hidden neuron. Note that $v_r = \lambda_0 \ubar{v}_r = 0$ and $\mb{w}_r = \ubar{\mb{w}}_r$. From \eqref{eq:pfThmICS_pert}, we have
\begin{subequations}
\begin{gather}
    \Delta \mb{t}_r = \alpha\cdot \sigma^\top\left((\ubar{\mb{w}}_r+\mb{u}_0)^\top X\right)  \\
    \Delta \mb{t}_i = \mb{0}, \quad i\in [m]\backslash\{r\}.
\end{gather}
\end{subequations}
Then, by decomposition \eqref{eq:decompose_empirical_loss2} we have
\begin{align}
\label{eq:pfThmICS_loss_diff}
    L(\theta')-L(\theta) &= \frac{1}{2}\left\|\Delta \mb{t}_r\right\|^2_2 + \left\langle \Delta \mb{y}, \Delta \mb{t}_r \right\rangle \nonumber \\
    & = \frac{1}{2}\alpha^2 \|\mb{p}\|^2_2 + \alpha \langle\Delta \mb{y}, \mb{p}\rangle
\end{align}
where we denote $\mb{p}=\sigma^\top((\ubar{\mb{w}}_r+\mb{u}_0)^\top X)\in \realvec{n}$ for brevity. Note that \eqref{eq:pfThmICS_loss_diff} is a quadratic function of $\alpha$, and by \eqref{eq:pfThmICS_nonzero} we have $\|\mb{p}\|_2, \langle\Delta \mb{y}, \mb{p}\rangle\not= 0$.

If $\langle\Delta \mb{y}, \mb{p}\rangle> 0$, we set
\begin{equation}
    \max\left\{-\frac{2\langle\Delta \mb{y}, \mb{p}\rangle}{\|\mb{p}\|^2_2}, -\sqrt{\delta^2-\|\mb{u}_0\|^2_2}\right\} < \alpha < 0
\end{equation}
and if $\langle\Delta \mb{y}, \mb{p}\rangle< 0$, we set
\begin{equation}
    0< \alpha < \min\left\{-\frac{2\langle\Delta \mb{y}, \mb{p}\rangle}{\|\mb{p}\|^2_2}, \sqrt{\delta^2-\|\mb{u}_0\|^2_2}\right\}.
\end{equation}
We simultaneously obtain $\theta' \in B(\theta, \delta)$ and
$L(\theta')<L(\theta)$. Because $\delta$ can be arbitrarily small, $\theta$ is not a local minimum of \widenet. By Proposition \ref{prop:stat_preserve}, $\theta$ is not a local maximum either. Thus, $\theta$ is a saddle point of \widenet. We obtain the first conclusion of the Theorem.

\subsubsection{All-Localmin Plateau}
Now, assume that the parent neuron is not locally effective but is locally constant. By Definition \ref{def:effective_neruon} and \ref{def:constant_neruon}, there exists a sufficiently small $\delta>0$ such that for any $\mb{u}\in B(\mb{0},\delta)$, we have
\begin{subequations}
\begin{gather}
\label{eq:pfThmICS_constant_z} 
    \sigma((\ubar{\mb{w}}_r+\mb{u})^\top X) =\sigma(\ubar{\mb{w}}_r X)\\
    \label{eq:pfThmICS_zero_inner_prod}
    \left\langle \Delta \mb{y},\sigma^\top((\ubar{\mb{w}}_r+\mb{u})^\top X) \right\rangle = \left\langle \Delta \mb{y},\sigma^\top(\ubar{\mb{w}}_r^\top X) \right\rangle = 0
\end{gather}
\end{subequations}
Let $\theta = \neusplit(\ubar{\theta},\pmb{\lambda})$ be an embedded point of \widenet with an arbitrary $\pmb{\lambda}\in \Lambda$. For any $\theta' \in B(\theta,\delta)$ we define an ``auxiliary'' perturbed point $\ubar{\theta}' = (\ubar{\mb{v}}',\ubar{W}') = (\ubar{\mb{v}}+\Delta \ubar{\mb{v}},\ubar{W}+\Delta \ubar{W})$ for \narrownet, given by
\begin{subequations}
\label{eq:pfThmICS_auxiliary_perturb}
\begin{gather}
    \Delta\ubar{\mb{w}}_i = \Delta \mb{w}_i,~~\Delta \ubar{v}_i = \Delta v_i,\quad i\in[r-1]\\
    \Delta\ubar{\mb{w}}_r = \mb{0},\quad \Delta \ubar{v}_r =\sum^{m}_{i'=r} \Delta v_{i'}.
\end{gather}
\end{subequations}
Clearly, we have $\|\Delta \ubar{W}\|_F \leq \|\Delta W\|_F$, By Cauchy–Schwarz inequality we have
\begin{equation}
    \|\Delta \ubar{\mb{v}}\|^2_2 = \sum^{r}_{i=1} \Delta \ubar{v}_i^2
        \leq
        \sum^{r-1}_{i=1} \Delta v_i^2 + (m-r+1)\sum^m_{i'=r}  \Delta v_{i'}^2  \leq  
        (m-r+1)\|\Delta \mb{v}\|_2^2.
\end{equation}
Denoting $C_1=\sqrt{m-r+1}$, we have $\ubar{\theta}'\in B(\ubar{\theta},C_1\delta)$ as long as $\theta' \in B(\theta,\delta)$.

For $i\in [r-1]$, we have
\begin{subequations}
\label{eq:pfThmICS_deltat}
\begin{align}
    \Delta \mb{t}_i^\top &= (v_i+\Delta v_i)\sigma\left((\mb{w}_i+\Delta \mb{w}_i)^\top X\right) - v_i\sigma(\mb{w}_i^\top X) \\
    &= (\ubar{v}_i+\Delta \ubar{v}_i)\sigma\left((\ubar{\mb{w}}_i+\Delta \ubar{\mb{w}}_i)^\top X\right)  - \ubar{v}_i \sigma(\ubar{\mb{w}}_i^\top X) \\
    & = \Delta \ubar{\mb{t}}_i^\top.
\end{align}
\end{subequations}
For $i\in[r:m]$, we have
\begin{subequations}
\begin{align}
    \Delta \mb{t}_i^\top &= (v_i+\Delta v_i)\sigma\left((\mb{w}_i+\Delta \mb{w}_i)^\top X\right) - v_i \sigma(\mb{w}_i^\top X)\\
    &= (\lambda_{i-r} \ubar{v}_r + \Delta v_i) \sigma\left((\ubar{\mb{w}}_r + \Delta \mb{w}_i)^\top X\right) - \lambda_{i-r}\ubar{v}_r \sigma(\ubar{\mb{w}}_r^\top X)\\
    &= \Delta v_i \sigma(\ubar{\mb{w}}_r^\top X)
\end{align}
\end{subequations}
where the last equality follows from \eqref{eq:pfThmICS_constant_z}. For $i \in [r:m]$, we have $\langle\Delta \mb{y}, \Delta \mb{t}_i \rangle = 0$ by \eqref{eq:pfThmICS_zero_inner_prod} . Further, from \eqref{eq:pfThmICS_zero_inner_prod} and \eqref{eq:pfThmICS_auxiliary_perturb} we have
\begin{subequations}
\label{eq:pfThmICS_deltat_r}
\begin{gather}
\Delta \ubar{\mb{t}}_r = \Delta \ubar{v}_r \sigma^\top(\ubar{\mb{w}}_r^\top X) = \sum^m_{i=r} \Delta v_i \sigma^\top(\ubar{\mb{w}}_r^\top X) \\ 
\langle \Delta \mb{y}, \Delta \ubar{\mb{t}}_r \rangle = \sum^m_{i=r} \Delta v_i \left\langle \Delta \mb{y}, \sigma^\top (\ubar{\mb{w}}_r^\top X) \right\rangle = 0.
\end{gather}
\end{subequations}

Combining \eqref{eq:pfThmICS_deltat}-\eqref{eq:pfThmICS_deltat_r} and decomposition \eqref{eq:decompose_empirical_loss2}, we have
\begin{subequations}
\label{eq:pfThmICS_loss_diff2}
\begin{align}
    L(\theta')-L(\theta) &= \frac{1}{2}\left\|\sum^m_{i=1} \Delta \mb{t}_i\right\|^2_2 + \sum^m_{i'=1} \left\langle \Delta \mb{y}, \Delta \mb{t}_{i'} \right\rangle\\
    & = \frac{1}{2}\left\|\sum^{r-1}_{i=1}\Delta \ubar{\mb{t}}_i + \sum^{m}_{i'=r} \Delta v_{i'} \sigma^\top(\ubar{\mb{w}}_r^\top X)\right\|_2^2+ \sum^{r-1}_{i''=1}\langle\Delta \mb{y}, \Delta \ubar{\mb{t}}_{i''}\rangle \\
    \label{eq:pfThmICS_loss_diff2_3}
    & = \frac{1}{2}\left\|\sum^{r}_{i=1}\Delta \ubar{\mb{t}}_i\right\|_2^2 + \sum^{r}_{i''=1}\langle\Delta \ubar{\mb{y}}, \Delta \ubar{\mb{t}}_{i''}\rangle \\
    & = \ubar{L}(\ubar{\theta}') - \ubar{L}(\ubar{\theta})
\end{align}
\end{subequations}
where \eqref{eq:pfThmICS_loss_diff2_3} utilizes the fact 
$\Delta \ubar{\mb{y}} = \ubar{\bhat{y}} - \mb{y} = \bhat{y} - \mb{y} = \Delta \mb{y}$ from Proposition \ref{prop:stat_preserve}. Because $\ubar{\theta}$ is a local minimum of \narrownet, for sufficiently small $\delta$ we have $\ubar{L}(\ubar{\theta}') - \ubar{L}(\ubar{\theta})\geq 0$ for any $\ubar{\theta}'\in B(\ubar{\theta},C_1 \delta)$. This yields $L(\theta')-L(\theta)\geq 0$ for any $\theta'\in B(\theta,\delta)$. Thus, $\theta$ is a local minimum of \widenet. We obtain the second conclusion of the Theorem.

\subsection{Proof of Theorem \ref{thm:splitting_saddle}}
Suppose that $\theta = (\ubar{\mb{v}},\ubar{W})$ is a saddle point of \narrownet. For any perturbed point $\ubar{\theta}' = (\ubar{\mb{v}}', \ubar{W}')$ of \narrownet, we construct the following perturbed point $\theta' = (\mb{v}', W')$ for \widenet:
\begin{subequations}
\label{eq:pfThmSS_perturb}
\begin{gather}
\label{eq:pfThmSS_perturb1}
   \Delta v_i = \Delta \ubar{v}_i, \quad \Delta \mb{w}_i = \Delta \ubar{\mb{w}}_i, \quad i\in[1:(r-1)] \\
    \label{eq::proof_case4_perturb2}
    \Delta v_{i'} = \lambda_{i'-r}\Delta\ubar{v}_r, \quad \Delta \mb{w}_{i'} = \Delta \ubar{\mb{w}}_r,~~ i'\in[r:m].
\end{gather}
\end{subequations}
By Cauchy–Schwarz inequality, we have
\begin{subequations}
\label{eq::proof_case4_perturbation_range}
    \begin{gather}
        \|\Delta \mb{v}\|^2_2 = \sum^{r-1}_{i=1} \Delta \ubar{v}_i^2 +  \sum^m_{i'=r}\lambda^2_{i'-r} \Delta \ubar{v}_r^2
        \leq \|\pmb{\lambda}\|^2_2\cdot  \|\Delta \ubar{\mb{v}}\|^2_2\\
        \|\Delta W\|_F^2 = 
        \sum^{r-1}_{i=1} \|\Delta\ubar{ \mb{w}}_i\|_2^2 + (m-r+1)\|\Delta\ubar{ \mb{w}}_r\|_2^2 \leq (m-r+1) \|\Delta \ubar{W}\|_F^2.
    \end{gather}
\end{subequations}
Denote $C=\max\{\sqrt{m-r+1}, \|\pmb{\lambda}\|_2\}$. Then, for any $\delta>0$, we have $\theta' \in B(\theta,\delta)$ as long as $\ubar{\theta}' \in B(\ubar{\theta}/C,\delta)$.

Next, we inspect the network output for $\theta'$. By Definition \ref{def:neuron_splitting} and  \eqref{eq:pfThmSS_perturb} we have
\begin{subequations}
\label{eq:pfThmSS_output}
\begin{align}
\hat{\mb{y}}'^\top &=\sum^{m}_{i=1}v'_i \sigma\left(\mb{w}'^\top_i X\right) \\
\label{eq:pfThmSS_output1}
& = \sum^{r-1}_{i=1} \ubar{v}'_i\sigma\left(\ubar{\mb{w}}'^\top_i X\right) + \sum^m_{i'=r} (\lambda_{i'-r} \ubar{v}_r + \Delta v_{i'})\sigma\left((\ubar{\mb{w}}_r +\Delta \mb{w}_{i'})^\top X\right) \\
\label{eq:pfThmSS_output2}
& = \sum^{r-1}_{i=1} \ubar{v}'_i\sigma\left(\ubar{\mb{w}}'^\top_i X\right) + \sum^m_{i'=r} \lambda_{i'-r} ( \ubar{v}_r + \Delta \ubar{v}_r)\sigma\left((\ubar{\mb{w}}_r +\Delta \ubar{\mb{w}}_r)^\top X\right) \\ 
& = \sum^{r-1}_{i=1} \ubar{v}'_i\sigma\left(\ubar{\mb{w}}'^\top_i X\right) +  (\ubar{v}_r + \Delta \ubar{v}_r)\sigma\left((\ubar{\mb{w}}_r +\Delta \ubar{\mb{w}}_r)^\top X\right)  \\ 
& = \sum^{r}_{i=1} (\ubar{v}_i + \Delta \ubar{v}_i)\sigma\left((\ubar{\mb{w}}_i +\Delta \ubar{\mb{w}}_i)^\top X\right) \\
& = \ubar{\hat{\mb{y}}}'^\top.
\end{align}
\end{subequations}
Because \narrownet and \widenet share the same training data $\mb{y}$, \eqref{eq:pfThmSS_output} implies
\begin{equation}
    L(\theta) = \frac{1}{2}\|\hat{\mb{y}}' - \mb{y} \|^2_2 =  \frac{1}{2}\|\ubar{\hat{\mb{y}}}' - \mb{y} \|^2_2 = \ubar{L}(\ubar{\theta}').
\end{equation}

Since $\theta$ is a saddle point of \narrownet, for any $\delta>0$ there exists $\ubar{\theta}' \in B(\ubar{\theta},\delta/C)$ such that $\ubar{L}(\ubar{\theta}')< \ubar{L}(\ubar{\theta})$. Then, we have $\theta' \in B(\theta,\delta)$ with 
\begin{equation}
    L(\theta')  = \ubar{L}(\ubar{\theta}') < \ubar{L}(\ubar{\theta}) = L(\theta)
\end{equation}
where the last inequality follows from Proposition \ref{prop:stat_preserve}. Therefore, we can find a perturbed point with a smaller empirical loss in an arbitrarily small neighborhood of $\theta$, and hence $\theta$ is not a local minimum of \widenet. From Lemma \ref{lem:no_localmax}, $\theta$ is not a local maximum either. Therefore, $\theta$ is a saddle point of \widenet. We complete the proof.

\bibliographystyle{unsrt}
\bibliography{references}

\appendix
\section{Proof of Lemmas and Propositions}
This appendix provides proofs of all the propositions and lemmas presented in the main body of this paper. 
\subsection{Proof of Proposition \ref{prop:stat_preserve}}
\label{app:pfPropSP}
By Definition \ref{def:neuron_splitting}, we have
\begin{equation}
    f(\mb{x};\theta) = \sum^m_{i=1}v_i\sigma(\mb{w}_i^\top \mb{x}) = \sum^{r-1}_{i=1}\ubar{v}_i\sigma(\ubar{\mb{w}}_i^\top \mb{x}) + \sum^{m}_{j=r}\lambda_{j-r}\ubar{v}_r\sigma(\ubar{\mb{w}}_r^\top \mb{x})
    = \sum^r_{i=1} \ubar{v}_i\sigma(\ubar{\mb{w}}_i^\top \mb{x}) = \ubar{f}(\mb{x};\ubar{\theta}).
\end{equation}
That is, \widenet has the same input-output mapping at $\theta$ with \narrownet at $\ubar{\theta}$. As the two networks share the same training data, they also yield the same prediction vector, i.e., $\bhat{y} = \ubar{\bhat{y}}$. Further, define the residual output $\Delta \mb{y} = \bhat{y}-\mb{y}$ and $\Delta \ubar{\mb{y}} = \ubar{\bhat{y}}-\mb{y}$ for \widenet and \narrownet, respectively. We have
$\Delta \mb{y} = \bhat{y}-\mb{y} = \ubar{\bhat{y}}-\mb{y} = \Delta \ubar{\mb{y}}$, yielding $L(\theta) = \ubar{L}(\ubar{\theta})$.

We next prove that $\theta$ is a stationary point of \widenet if and only if $\ubar{\theta}$ of \narrownet. To this end, we establish a relationship between the gradients of $L$ at $\theta = (\mb{v},W)$ and the gradients of $\ubar{L}$ at $\ubar{\theta} = (\ubar{\mb{v}},\ubar{W})$. For the gradients with respect to the output weights, we have
\begin{subequations}
\label{eq:pfPropSP_gradient_v}
\begin{gather}
\label{eq:pfPropSP_gradient_v1}
    \frac{\partial L}{\partial v_i} = \langle \Delta \mb{y}, \mb{z}_i\rangle =  \langle \Delta \ubar{\mb{y}}, \ubar{\mb{z}}_i\rangle = \frac{\partial \ubar{L}}{\partial \ubar{v}_i},\quad i\in[r-1]  \\
\label{eq:pfPropSP_gradient_v2}
    \frac{\partial L}{\partial v_{i'}} = \langle \Delta \mb{y}, \mb{z}_{i'} \rangle
    =  \langle \Delta \ubar{\mb{y}}, \ubar{\mb{z}}_r\rangle = \frac{\partial \ubar{L}}{\partial \ubar{v}_r},\quad i'\in [r:m]
\end{gather}
\end{subequations}
where \eqref{eq:pfPropSP_gradient_v1} follows from $\mb{w}_i = \ubar{\mb{w}}_i$ for $i\in[r-1]$, and \eqref{eq:pfPropSP_gradient_v2} follows from $\mb{w}_{i'}= \ubar{\mb{w}}_r$ for $i'\in[r:m]$.
Note that \eqref{eq:pfPropSP_gradient_v} is regardless of the choice of $\pmb{\lambda}$. Thus, $\partial L/\partial \mb{v} = \mb{0}$ if and only if $\partial \ubar{L}/\partial \ubar{\mb{v}} = \mb{0}$.

Next, we consider the gradients with respect to the hidden weights. For \narrownet, we have
\begin{subequations}
\begin{gather}
    \frac{\partial \ubar{L}}{\partial \ubar{\mb{w}}_i} = \ubar{v}_i
    \begin{bmatrix}
    \langle \Delta \ubar{\mb{y}}, \ubar{\mb{z}}_i' \circ \mb{x}_1 \rangle \\
    \langle \Delta \ubar{\mb{y}}, \ubar{\mb{z}}_i' \circ \mb{x}_2 \rangle \\
    \vdots \\
    \langle \Delta \ubar{\mb{y}}, \ubar{\mb{z}}_i' \circ \mb{x}_d \rangle 
    \end{bmatrix}, ~~ i\in[r].
\end{gather}
\end{subequations}
For \widenet, we have
\begin{subequations}
\begin{gather}
\label{eq:pfPropSP_gradient_w1}
    \frac{\partial L}{\partial \mb{w}_i} = v_i 
    \begin{bmatrix}
    \langle \Delta \mb{y}, \mb{z}_i' \circ \mb{x}_1 \rangle \\
    \langle \Delta \mb{y}, \mb{z}_i' \circ \mb{x}_2 \rangle \\
    \vdots \\
    \langle \Delta \mb{y}, \mb{z}_i' \circ \mb{x}_d \rangle 
    \end{bmatrix} = \ubar{v}_i
    \begin{bmatrix}
    \langle \Delta \ubar{\mb{y}}, \ubar{\mb{z}}_i' \circ \mb{x}_1 \rangle \\
    \langle \Delta \ubar{\mb{y}}, \ubar{\mb{z}}_i' \circ \mb{x}_2 \rangle \\
    \vdots \\
    \langle \Delta \ubar{\mb{y}}, \ubar{\mb{z}}_i' \circ \mb{x}_d \rangle 
    \end{bmatrix} = \frac{\partial\ubar{E}}{\partial \ubar{\mb{w}}_i},~~ i\in [r-1]\\
    %line2
\label{eq:pfPropSP_gradient_w2}
    \frac{\partial L}{\partial \mb{w}_{i'}} = v_{i'} 
    \begin{bmatrix}
    \langle \Delta \mb{y}, \mb{z}_{i'}' \circ \mb{x}_1 \rangle \\
    \langle \Delta \mb{y}, \mb{z}_{i'}' \circ \mb{x}_2 \rangle \\
    \vdots \\
    \langle \Delta \mb{y}, \mb{z}_{i'}' \circ \mb{x}_d \rangle 
    \end{bmatrix} = \lambda_{i'-r}\ubar{v}_r
    \begin{bmatrix}
    \langle \Delta \ubar{\mb{y}}, \ubar{\mb{z}}_r' \circ \mb{x}_1 \rangle \\
    \langle \Delta \ubar{\mb{y}}, \ubar{\mb{z}}_r' \circ \mb{x}_2 \rangle \\
    \vdots \\
    \langle \Delta \ubar{\mb{y}}, \ubar{\mb{z}}_r' \circ \mb{x}_d \rangle 
    \end{bmatrix} = \lambda_{i'-r} \frac{\partial\ubar{L}}{\partial \ubar{\mb{w}}_r}, ~~ i' \in [r:m].
\end{gather}
\end{subequations}
If $\partial \ubar{L}/\partial \ubar{W} = \mb{0}$, we readily have $\partial L/\partial W = \mb{0}$. Conversely, suppose that $\partial L/\partial W = \mb{0}$. From \eqref{eq:pfPropSP_gradient_w1}, $\partial \ubar{L}/\partial \ubar{\mb{w}}_i = \mb{0}$ for all $i\in [r-1]$. Further, noting that $\sum^{m-r}_{j=0}\lambda_j = 1$ for any $\pmb{\lambda}\in \Lambda$, there exists $i_0 \in [r:m]$ such that $\lambda_{i_0-r}> 0$. Then, from \eqref{eq:pfPropSP_gradient_w2} we have \begin{equation}
    \frac{\partial \ubar{L}}{\partial \ubar{\mb{w}}_r} = \frac{1}{\lambda_{i_0-r}} \cdot \frac{\partial \ubar{L}}{\partial \mb{w}_{i_0}} = \mb{0},
\end{equation}
yielding $\partial \ubar{L}/\partial \ubar{W} = \mb{0}$. We conclude that $\partial L /\partial W = \mb{0}$ if and only if $\partial \ubar{L} /\partial \ubar{W} = \mb{0}$.

Therefore, for any $\pmb{\lambda} \in \Lambda$, $\theta$ is a stationary point of \widenet if and only if $\ubar{\theta}$
is a stationary point of \narrownet. We complete the proof.

\subsection{Proof of Proposition \ref{prop:condition-connection}}\label{app:pfPropCondC}

We first prove that the $i$-th neuron is locally effective if $\Hin_i\not=0$. Equivalently, we prove that $\Hin_i=0$ if the $i$-th neuron is locally ineffective. Assume that the $i$-th neuron is locally ineffective. Then there exists $\delta>0$ such that  
\begin{equation}\label{eq:cond_locally_ineff}
\left \langle \bhat{y}- \mb{y}, \sigma^\top(\mb{w}'^{\top}_i X)\right \rangle = 0, \quad \forall \mb{w}'_i\in B(\mb{w}_i, \delta).
\end{equation}

Denote $F_i(\mb{u})=\left \langle \bhat{y}- \mb{y}, \sigma^\top(\mb{u}^{\top} X)\right \rangle$. By \eqref{eq:cond_locally_ineff}, we obtain that $F_i(\mb{w}'_i)\equiv0$ for all $\mb{w}'_i\in B(\mb{w}_i, \delta)$. This implies that for all $j, j'\in [d]$, the partial second-order derivative $\partial^2_{j,j'}F_i(\mb{w}_i)=0$. Therefore,
\begin{equation*}
    (\Hin_i)_{j,j'} = v_i\langle \bhat{y} -\mb{y}, \mb{z}_i'' \circ \mb{x}_j \circ \mb{x}_{j'} \rangle=v_i\partial^2_{j,j'}F_i(\mb{w}_i) = 0, \quad \forall j, j'\in [d],
\end{equation*}
which implies $\Hin_i=0$. The proof is complete.

Next we prove that if the $i$-th neuron of the network is locally constant, then it is not locally effective. Assume that the $i$-th neuron of the network is locally constant, then there exists $\delta>0$ such that
\begin{equation*}
\sigma(\mb{w}'^{\top}_i X) = \sigma(\mb{w}^{\top}_i X),\quad \forall \mb{w}_i' \in B(\mb{w}_i, \delta).
\end{equation*}
Notice that $(W, \mb{v})$ is a stationary point. Therefore, we have
\begin{equation}\label{eq:cond_partial_vi}
    0=\frac{\partial L}{\partial v_i}=\left \langle \bhat{y}- \mb{y}, \sigma^\top(\mb{w}^{\top}_i X)\right \rangle.
\end{equation}
Combining with \eqref{eq:cond_partial_vi}, we obtain
\begin{equation*}
    \left \langle \bhat{y}- \mb{y}, \sigma^\top(\mb{w}'^{\top}_i X)\right \rangle = 0, \quad \forall \mb{w}'_i\in B(\mb{w}_i, \delta),
\end{equation*}
implying that the $i$-th neuron is locally ineffective. We complete the proof.

\subsection{Proof of Lemma \ref{lem:no_localmax}}
\label{app:pfLemNL}
If the empirical loss $L$ is locally constant at $\theta$, $\theta$ is a local-max of $L$. In the following we prove the converse. That is, if $\theta$ is a local-max of $L$, there exists $\delta>0$ such that
\begin{equation}
\label{eq:pfLemNL_objective}
    L(\theta') = L(\theta), ~~\forall \theta' \in B(\theta,\delta). 
\end{equation}

First, suppose that there exists $i_0\in[m]$ such that $\mb{z}_{i_0} \not= \mb{0}$. Without loss of generality we can assume $i_0 = 1$. For an arbitrary $\delta>0$, we construct a perturbed point $\theta' = (\mb{v}', W')$ as
\begin{equation}
    \label{eq:pfLemNL_pert}
    v_1' = v_1 + \alpha, \quad  v_i' = v_i, \quad \forall i\in[2:m], \quad W' = W 
\end{equation}
where $\alpha \in \mathbb{R}$ with $0<|\alpha|<\delta$, whose sign will be determined later. Clearly, $\theta' \in B(\theta,\delta)$. By the decomposition \eqref{eq:decompose_empirical_loss2} we have
\begin{equation}
\label{eq:pfLemNL_loss_diff}
    L(\theta')-L(\theta) = \frac{1}{2}\left\|\sum^m_{i=1} \Delta \mb{t}_i\right\|^2_2 + \left\langle \Delta \mb{y}, \sum^m_{i'=1} \Delta \mb{t}_{i'} \right\rangle =\frac{1}{2}\alpha^2 \|\mb{z}_1\|^2_2 + \alpha \langle \Delta \mb{y}, \mb{z}_1\rangle.
\end{equation}
Recall that we assume $\mb{z}_1 \not= \mb{0}$. Set $\alpha > 0$ if $\langle \Delta \mb{y}, \mb{z}_1\rangle \geq 0$, and $\alpha <0$ if $\langle \Delta \mb{y}, \mb{z}_1\rangle < 0$. This yields $L(\theta')-L(\theta) > 0$. Since $\delta$ can be arbitrarily small, a contradiction to the assumption that $\theta$ is a local maximum of $L$. Thus, we must have $\mb{z}_i = \mb{0}$ for all $i\in[m]$. 

Next, suppose that for any $\delta>0$, there exists $i_1\in[m]$ and $\mb{u} \in\realvec{d}$ with $0<\|\mb{u}\|_2 < \delta$, such that $\sigma((\mb{w}_{i_1}+\mb{u})^\top X) \not=\mb{0}^\top$. Without loss of generality we assume $i_1 = 1$. Consider the following perturbation:
\begin{subequations}
\begin{gather}
    v'_1 = v_1 + \beta, \quad  \mb{w}'_1 = \mb{w}_1 + \mb{u}\\
    v'_i = v_i, \quad \mb{w}'_i = \mb{w}_i, \quad \forall i\in[2:m] 
\end{gather}
\end{subequations}
where $\beta \in \mathbb{R}$ with $0<|\beta|^2<\delta^2-\|\mb{u}\|^2_2$, whose sign will be determined later. Clearly, $\theta' \in B(\theta,\delta)$. Similar to \eqref{eq:pfLemNL_loss_diff}, we have
\begin{equation}
    L(\theta')-L(\theta) =\frac{1}{2}\beta^2\|\Delta \mb{z}_1\|^2_2 + \beta \langle \Delta \mb{y}, \Delta\mb{z}_1\rangle 
\end{equation}
where we denote
\begin{equation}
    \Delta \mb{z}_1^\top = \sigma((\mb{w}_1+\mb{u})^\top X) - \mb{z}_1^\top = \sigma((\mb{w}_1+\mb{u})^\top X) \not= \mb{0}^\top.
\end{equation}
We set $\beta > 0$ if $\langle \Delta \mb{y}, \Delta \mb{z}_1\rangle \geq 0$, and $\beta <0$ if $\langle \Delta \mb{y}, \Delta  \mb{z}_1\rangle < 0$. This yields $L(\theta')-L(\theta) > 0$. Since $\delta$ can be arbitrarily small, a contradiction to the assumption that $\theta$ is a local maximum. Thus, there exists $\delta>0$ such that $\sigma(\mb{w}_i'^\top X) = \mb{0}^\top$ for any $i\in [m]$ and $\mb{w}_i' \in B(\mb{w}_i, \delta)$. As a result, the network outputs $\bhat{y} = \mb{0}$ locally at $\theta$, and hence $L$ is locally constant at $\theta$. We complete the proof.

\subsection{Proof of Lemma \ref{lem:PD_localmin}}
\label{app:pfLemPDL}
By the definition of $\Delta \btil{w}_i$ in \eqref{eq:pfThmSL_PD_localmin_def_tilw}, we have
\begin{equation}
\sum^m_{i=r}\lambda_{i-r}'\Delta \btil{w}_i = \sum^m_{i=r}\lambda_{i-r}'\left(\Delta \mb{w}_i - \Delta \ubar{\mb{w}}_r\right) = \sum^m_{i=r}\lambda_{i-r}'\Delta \mb{w}_i - \Delta \ubar{\mb{w}}_r = \mb{0}
\end{equation}
where the last equality follows from \eqref{eq:pfThmSL_PD_localmin_auxiliary_perturb}. This implies
\begin{equation}
\ubar{v}_r'\sigma'\left(\ubar{\mb{w}}_r'^\top \mb{x}\upbrac{k}\right)\left(\sum^m_{i=r}\lambda_{i-r}'\Delta\btil{w}_i^\top\mb{x}\upbrac{k}\right) = 0.
\end{equation}
where $\mb{x}\upbrac{k}\in \realvec{d}$ is the $k$-th training input, i.e., the $k$-th column of $X$, $k\in[n]$. Subtracting this zero term to the expression \eqref{eq:pfThmSL_PD_localmin_def_e}, we re-write the $k$-th entry of $\mb{e}$ as
\begin{subequations}
\label{eq:pfLemPDL_ek}
\begin{align}
\label{eq:pfLemPDL_ek_1}
    e_k  &= \sum_{i=r}^m v_i' \sigma\left(\mb{w}_i'^\top \mb{x}\upbrac{k}\right) -  \ubar{v}'_r\sigma\left(\ubar{\mb{w}}_r'^\top\mb{x}\upbrac{k}\right) - \ubar{v}'_r \sigma'\left(\ubar{\mb{w}}_r'^\top\mb{x}\upbrac{k}\right)\left(\sum^m_{i'=r}\lambda_{i'-r}'\Delta \btil{w}_{i'}^\top\mb{x}\upbrac{k}\right) \\
    \label{eq:pfLemPDL_ek_2}
    & = \sum^m_{i=r}\lambda_{i-r}' \ubar{v}_r'\left[\sigma\left((\ubar{\mb{w}}_r' + \Delta \btil{w}_i)^\top \mb{x}\upbrac{k}\right) - \sigma\left(\ubar{\mb{w}}_r'^\top\mb{x}\upbrac{k}  \right) - \sigma'\left(\ubar{\mb{w}}_r'^\top\mb{x}\upbrac{k}\right)\Delta \btil{w}_i^\top\mb{x}\upbrac{k}\right]
\end{align}
\end{subequations}
where \eqref{eq:pfLemPDL_ek_2} follows from $v_i' = \lambda'_{i-r} \ubar{v}'_r$. Define $F_{i,k}:[0,1] \rightarrow \mathbb{R} $ as
\begin{equation}
F_{i,k}(t) = \sigma\left((\ubar{\mb{w}}_r' + t \cdot \Delta \btil{w}_i)^\top \mb{x}\upbrac{k}\right) -  \sigma\left(\ubar{\mb{w}}_r'^\top\mb{x}\upbrac{k}\right) - t \sigma'\left(\ubar{\mb{w}}_r'^\top\mb{x}\upbrac{k}\right)\Delta \btil{w}_i^\top\mb{x}\upbrac{k}.
\end{equation}
Clearly, this is a twice continuous differentiable function with $F_{i,k}(0) = F'_{i,k}(0) = 0$, and
\begin{equation}
\label{eq:pfLemPDL_Fpp}
    F''_{i,k}(t) = \sigma''\left((\ubar{\mb{w}}_r' + t\cdot  \Delta \btil{w}_i)^\top \mb{x}\upbrac{k}\right)\cdot\left(\Delta \btil{w}_i^\top \mb{x}\upbrac{k}\right)^2,\quad i\in[r:m],\quad k\in[n].
\end{equation}
We define
\begin{equation}
\label{eq:pfLemPDL_def_big_M}
    M(\delta)\triangleq\max_{\substack{\theta'\in \bar{B}(\theta,\delta)\\ i\in[r:m],k\in[n]\\ t\in[0,1]}} \left|\sigma''\left((\ubar{\mb{w}}_r' + t \cdot \Delta \btil{w}_i)^\top \mb{x}\upbrac{k}\right)\right|\cdot \left\|\mb{x}\upbrac{k}\right\|_2^2
\end{equation}
where $\bar{B}(\theta,\delta)$ denotes the closure of $B(\theta,\delta)$, which is a compact set. Thus, the maximum in \eqref{eq:pfLemPDL_def_big_M} can be achieved, and hence $M(\delta)$ is well defined as a finite non-negative value for any $\delta >0$. Also, given a $\delta>0$, $M(\delta)$ is independent of $\Delta \mb{v}$, $\Delta W$, and $t$. Combining \eqref{eq:pfLemPDL_Fpp} and \eqref{eq:pfLemPDL_def_big_M}, we have $|F''_{i,k}(t)| \leq M(\delta) \|\Delta \btil{w}_i\|^2_2$. Then, combined with \eqref{eq:pfLemPDL_ek} we have
\begin{align}
    |e_k| &=  \left|\sum^m_{i=r}\lambda_{i-r}'\ubar{v}_r'\int_{0}^1\int_{0}^t F''_{i,k}(\tau)d\tau dt\right| \nonumber \\
    &\leq \sum^m_{i=r}\lambda_{i-r}' |\ubar{v}_r'| \int_{0}^1\int_{0}^t  \|\Delta \btil{w}_i\|^2_2 M(\delta) d\tau dt   \nonumber \\
    &\leq  \frac{1}{2}|\ubar{v}_r'|\cdot M(\delta) \sum^m_{i=r}\lambda_{i-r}'\|\Delta \btil{w}_i\|^2_2.
\end{align}
There exists sufficiently small $\delta_{1,1}>0$ such that $\lambda_{i-r}' \leq 2\lambda_{i-r}$ and $|\ubar{v}'_r| \leq 2|\ubar{v}_r|$ for all $\theta' \in B(\theta,\delta_{1,1})$. Thus, for any $0 <\delta \leq \delta_{1,1}$, we have
\begin{equation}
    |e_k| \leq 2|\ubar{v}_r|\cdot M(\delta) \sum^m_{i=r}\lambda_{i-r}\|\Delta \btil{w}_i\|^2_2, \quad \forall k \in [n], \quad \forall \theta'\in B(\theta, \delta),
\end{equation}
implying
\begin{equation}
\label{eq:pfLemPDL_bound1_cond1}
    \|\mb{e}\|_2 = \sqrt{\sum^n_{k=1}e_k^2}
    \leq 2 \sqrt{n}\cdot |\ubar{v}_r|\cdot M(\delta) \sum^m_{i=r}\lambda_{i-r}\|\Delta \btil{w}_k\|^2_2, \quad \forall \theta' \in B(\theta, \delta).
\end{equation}

Next, because $\ubar{\bhat{y}}$ is a continuous function with respect to $\theta$, for any $\epsilon >0$ there exists a sufficiently small $\delta_{1,2}>0$ such that
\begin{equation}
\label{eq:pfLemPDL_bound1_cond2}
    \left\|\ubar{\bhat{y}}'-\ubar{\bhat{y}}\right\|_2 \leq \frac{\epsilon}{4\sqrt{n}\cdot|\ubar{v}_r|\cdot M(1)},\quad  \forall \theta' \in B(\theta,\delta_{1,2}).
\end{equation}
Let $\delta_1 = \min\{\delta_{1,1},\delta_{1,2},1\}$, with  \eqref{eq:pfLemPDL_bound1_cond1} and \eqref{eq:pfLemPDL_bound1_cond2} we have that for any $\theta' \in B(\theta,\delta_1)$,
\begin{subequations}
\label{eq:pfLemPDL_ineq1}
\begin{align}
    \left\|(\ubar{\bhat{y}}'-\ubar{\bhat{y}}) + \mb{e}\right\|^2_2  &= \left\|\ubar{\bhat{y}}'-\ubar{\bhat{y}}\right\|^2_2 + 2\mb{e}^\top (\ubar{\bhat{y}}'-\ubar{\bhat{y}}) + \|\mb{e}\|^2_2 \\
    &\geq \left\|\ubar{\bhat{y}}'-\ubar{\bhat{y}}\right\|^2_2 - 2\|\mb{e}\|_2\cdot \left\|\ubar{\bhat{y}}'-\ubar{\bhat{y}}\right\|_2 \\
    \label{eq:pfLemPDL_ineq1_1}
    &\geq \left\|\ubar{\bhat{y}}'-\ubar{\bhat{y}}\right\|_2^2 - \frac{M(\delta_1)}{M(1)}\cdot \epsilon \sum^m_{i=r}\lambda_{i-r}\|\Delta \btil{w}_i\|^2_2 \\
    \label{eq:pfLemPDL_ineq1_2}
    &\geq \left\|\ubar{\bhat{y}}'-\ubar{\bhat{y}}\right\|_2^2 - \epsilon \sum^m_{i=r}\lambda_{i-r}\|\Delta \btil{w}_i\|^2_2
\end{align}
\end{subequations}
where \eqref{eq:pfLemPDL_ineq1_2} follows from the fact that the function $M$ is monotonically increasing by its definition \eqref{eq:pfLemPDL_def_big_M}. We obtain the first inequality of the lemma.

In the sequel we bound $\langle\Delta \mb{y}, \mb{e}\rangle$ in terms of $\sum^m_{i=r}\lambda_{i-r}\|\Delta \btil{w}_i\|^2_2$. By \eqref{eq:pfThmSL_PD_localmin_auxiliary_perturb} and \eqref{eq:pfThmSL_PD_localmin_def_e} we have
\begin{align}
\label{eq:pfLemPDL_bound2_cond}
\langle \Delta \mb{y}, \mb{e}\rangle &= ~\left\langle \Delta \mb{y}, \sum^m_{i=r}v_i' \sigma\left(\sum^d_{k=1}w_{i,k}'\mb{x}_k\right) - \ubar{v}_r'\sigma\left(\sum^d_{k=1}\ubar{w}_{r,k}'\mb{x}_k\right)\right\rangle \nonumber\\
&=\sum^m_{i=r}\lambda'_{i-r}\ubar{v}_r'\left\langle\Delta \mb{y}, \sigma\left(\sum^d_{k=1}(w_{r,k}+\Delta w_{i,k})\mb{x}_k\right)\right\rangle - \ubar{v}_r'\left\langle\Delta \mb{y}, \sigma\left(\sum^d_{k=1}(w_{r,k}+\Delta \ubar{w}_{r,k})\mb{x}_k\right) \right\rangle
\end{align}
where $\mb{x}_k \in \realmat{n}{1}$ denotes (the transpose of) the $k$-th row of $X$. We define a function $G_a: \realvec{d}\rightarrow \mathbb{R}$ parameterized by $a\in\mathbb{R}$ as
\begin{equation}
    G_a(\mb{u}) = a \left\langle \Delta \mb{y}, \sigma\left(\sum^d_{k=1}(w_{r,k}+u_k) \mb{x}_k\right)\right\rangle.
\end{equation}
Then, we can rewrite \eqref{eq:pfLemPDL_bound2_cond} as
\begin{equation}
\label{eq:pfLemPDL_bound2_cond_inG}
\langle \Delta \mb{y}, \mb{e}\rangle= \sum^m_{j=r}\lambda_{j-r}'G_{\ubar{v}_r'}(\Delta \mb{w}_j) - G_{\ubar{v}_r'}(\Delta \ubar{\mb{w}}_r).
\end{equation}
For a given $a\in \mathbb{R}$, let $H_{G}(\mb{u}; a) \in \realmat{d}{d}$ denotes the Hessian matrix of $G_a(\mb{u})$ with respect to $\mb{u}$, whose entries are given by
\begin{equation}
    \left(H_{G}(\mb{u};a)\right)_{j,j'} =  a \left\langle \Delta \mb{y}, \sigma''\left(\sum^d_{k=1}(w_{r,k}+u_k)\mb{x}_k\right)\circ \mb{x}_j\circ \mb{x}_{j'}\right\rangle, ~~ j,j'\in[d].
\end{equation}
By Proposition \ref{prop:stat_preserve} we have $\Delta \ubar{\mb{y}} = \Delta \mb{y}$, yielding
\begin{subequations}
\begin{align}
\left(H_G(\mb{0};\ubar{v}_r)\right)_{j,j'} &= \ubar{v}_r \left\langle \Delta \ubar{\mb{y}}, \sigma''\left(\sum_{k=1}^d w_{r,k}\cdot \mb{x}_k\right)\circ \mb{x}_j \circ \mb{x}_{j'} \right\rangle  \\
\label{eq:pfLemPDL_HG_2}
& = \ubar{v}_r \left\langle \Delta \ubar{\mb{y}}, \sigma''\left(\sum_{k=1}^d \ubar{w}_{r,k}\cdot \mb{x}_k\right)\circ \mb{x}_j \circ \mb{x}_{j'} \right\rangle \\
\label{eq:pfLemPDL_HG_3}
&= \ubar{v}_r \left\langle \Delta \ubar{\mb{y}}, \ubar{\mb{z}}''_r \circ \mb{x}_j \circ \mb{x}_{j'} \right\rangle = (\ubarHin_r)_{j,j'}
\end{align}
\end{subequations}
where \eqref{eq:pfLemPDL_HG_2} follows from $\ubar{\mb{w}}_r = \mb{w}_r$ and \eqref{eq:pfLemPDL_HG_3} follows from the definition of inner Hessian in \eqref{eq:def_inner_Hessian}. Thus, we have $H_{G}(\mb{0}; \ubar{v}_r) = \ubarHin_r$ which is positive definite by assumption. 

Note that $H_{G}(\mb{u}; a)$ is a continuous function of $(\mb{u},a)$. Thus, there exists a sufficiently small $\delta_{2,1}>0$ such that for any $\theta'\in B(\theta,\delta_{2,1})$ and $\mb{u}\in  B(\mb{0}, C_1 \delta_{2,1})$,\footnote{Recall that the constant $C_1=\sqrt{m-r+1}$. From \eqref{eq:pfThmSL_PD_localmin_perturb_range}, we have $\ubar{\theta}' \in B(\ubar{\theta},C_1\delta_{2,1})$ as long as $\theta' \in B(\theta,\delta_{2,1})$. This yields $\|\Delta \mb{w}_{r}\|_2, \|\Delta \mb{w}_{r+1}\|_2, \|\Delta \ubar{\mb{w}}_r\|_2 < C_1\delta_{2,1}$.} we have
\begin{equation}
    \lambda_{\min}\left(H_G(\mb{u};\ubar{v}_r')\right) > \frac{1}{2}\lambda_{\min}\left(H_G(\mb{0};\ubar{v}_r)\right) = \frac{1}{2}\lambda_{\min}(\ubarHin_r) >0.
\end{equation}
Thus, $G_{\ubar{v}_r'}(\mb{u}) - \lambda_{\min}(\ubarHin_r)\left\|\mb{u}\right\|^2_2/2$ is a convex function of $\mb{u}$ for $\mb{u} \in B(\mb{0}, C_1\delta_{2,1})$. By Jensen's inequality, we have
\begin{equation}
\label{eq:pfLemPDL_bound2_cond2}
    \sum^m_{i=r}\lambda_{i-r}'\left(G_{\ubar{v}_r'}(\Delta \mb{w}_i)-\frac{\lambda_{\min}(\ubarHin_r)}{2}\|\Delta \mb{w}_i\|^2_2\right)
    \geq G_{\ubar{v}_r'}\left(\sum^m_{i=r}\lambda_{i-r}'\Delta \mb{w}_i\right) - \frac{\lambda_{\min}(\ubarHin_r)}{2}\left\|\sum^m_{i=r}\lambda_{i-r}'\Delta \mb{w}_i\right\|^2_2.
\end{equation}
Noting that $\Delta \ubar{\mb{w}}_r = \sum^m_{i=r}\lambda_{i-r}'\Delta \mb{w}_i$, we combine \eqref{eq:pfLemPDL_bound2_cond_inG} and \eqref{eq:pfLemPDL_bound2_cond2}, yielding
\begin{align}
    \langle \Delta \mb{y}, \mb{e}\rangle & \geq \frac{1}{2}\lambda_{\min}(\ubarHin_r)
    \left(\sum^m_{i=r}\lambda_{i-r}'\|\Delta \mb{w}_i\|^2_2
    - \|\Delta \ubar{\mb{w}}_r\|^2_2
    \right) \nonumber \\
    & = \frac{1}{2}\lambda_{\min}(\ubarHin_r)
    \left(\sum^m_{i=r}\lambda_{i-r}'\|\Delta \mb{w}_i\|^2_2
    - 2\|\Delta \ubar{\mb{w}}_r\|^2_2 + \|\Delta \ubar{\mb{w}}_r\|^2_2
    \right) \nonumber \\
    & = \frac{1}{2}\lambda_{\min}(\ubarHin_r)
    \left(\sum^m_{i=r}\lambda_{i-r}'\|\Delta \mb{w}_i\|^2_2
    - 2\left\langle \sum^m_{i'=r} \lambda_{i'-r}' \Delta \mb{w}_{i'}, \Delta \ubar{\mb{w}}_r \right\rangle + \|\Delta \ubar{\mb{w}}_r\|^2_2
    \right) \nonumber \\
    & = \frac{1}{2}\lambda_{\min}(\ubarHin_r)
    \sum^m_{i=r}\lambda_{i-r}'\|\Delta \mb{w}_i - \Delta \ubar{\mb{w}}_r\|^2_2 \nonumber \\
    & = \frac{1}{2}\lambda_{\min}(\ubarHin_r)\sum^m_{i=r}\lambda_{i-r}'\|\Delta \btil{w}_i\|^2_2.
\end{align}
Further, there exists a sufficiently small $\delta_{2,2}>0$ such that $\lambda_{i-r}'\geq \lambda_{i-r}/2$ for all $\theta' \in B(\theta,\delta_{2,2})$, $i\in[r:m]$. Letting $\delta_2 = \min\{\delta_{2,1},\delta_{2,2}\}$, we have
\begin{equation}
\label{eq:pfLemPDL_ineq2}
\langle \Delta \mb{y}, \mb{e}\rangle \geq \frac{1}{4}\lambda_{\min}(\ubarHin_r)\sum^m_{i=r}\lambda_{i-r}\|\Delta \btil{w}_i\|^2_2.
\end{equation}
for all $\theta' \in B(\theta, \delta_2)$. We obtain the second inequality of the lemma.

Finally, setting $\delta = \min\{\delta_1, \delta_2\}$, both \eqref{eq:pfLemPDL_ineq1} and \eqref{eq:pfLemPDL_ineq2} hold. We complete the proof of Lemma \ref{lem:PD_localmin}.

\subsection{Proof of Lemma \ref{lem:ND_localmin}}
\label{app:pfLemNDL}
As in Section \ref{sect:pfThmSL_ND_localmin}, we assume $\lambda_0>0$ and $\lambda_1<0$ without loss of generality. By the definition \eqref{eq:pfThmSL_ND_localmin_def_ub}, we have $0<\beta_0,\beta_1<1$ and $\beta_0+\beta_1 = 1$. Define ``perturbed'' versions of $\beta_0, \beta_1$ as
\begin{equation}
\label{eq:pfLemNDL_def_lambdaprime}
    \beta'_0 \triangleq \frac{1}{\lambda_0'}, \quad \beta_1' \triangleq \frac{|\lambda_1'|}{\lambda_0'}
\end{equation}
where $\lambda'_0, \lambda'_1$ are given by \eqref{eq:pfThmSL_ND_localmin_lamprime}. We have $\beta_0'+\beta_1'=1$. Note that we assume a sufficiently small $\delta$ such that $|\lambda_j|/2 < |\lambda_j'| < 2|\lambda_j|$, and thus $0<\beta_j/4<\beta_j'<4\beta_j$, $j \in \{0,1\}$. 

Next, we rewrite $\ubar{v}_r'$, $v_{r+1}'$ in terms of $\beta_0'$, $\beta_1'$, and $v_{r}'$. Combining \eqref{eq:pfThmSL_ND_localmin_def_ub} and \eqref{eq:pfLemNDL_def_lambdaprime}, we have
\begin{subequations}
\label{eq:pfLemNDL_transfer}
\begin{equation}
    \ubar{v}_{r}' = \frac{1}{\lambda_0'}v_r' = \beta_0'v_r', \quad v_{r+1}'=\lambda_1 \ubar{v}_r' = -\frac{|\lambda_1|}{\lambda_0}v_r' = -\beta_1' v_r'.
\end{equation}
Similarly, from \eqref{eq:pfThmSL_ND_localmin_def_ub} and the fact that $\mb{w}_r = \mb{w}_{r+1} = \ubar{\mb{w}}_r$, we have
\begin{gather}
    \ubar{\mb{w}}_r' = \ubar{\mb{w}}_r + \Delta \ubar{\mb{w}}_r = \mb{w}_r + \Delta \mb{w}_r + \mb{u}_0 = \mb{w}_r' +\mb{u}_0 \\
    \mb{w}_{r+1}' = \mb{w}_{r+1} + \Delta \mb{w}_{r+1} = \mb{w}_r + \Delta \mb{w}_r + \mb{u}_1 = \mb{w}_r' +\mb{u}_1.
\end{gather}
\end{subequations}
The expressions in \eqref{eq:pfLemNDL_transfer} allow us to rewrite \eqref{eq:pfThmSL_ND_localmin_def_e} as
\begin{align}
\label{eq:pfLemNDL_rewrite_e}
    \mb{e}^\top &=
    v_r'\sigma(\mb{w}_r'^\top X) - \ubar{v}_r'\sigma(\ubar{\mb{w}}_r'^\top X) + v_{r+1}'\sigma(\mb{w}_{r+1}'^\top X)  \nonumber \\
    &= v_r'\sigma(\mb{w}_r'^\top X) - \beta_0' v_r'  \sigma\left((\mb{w}_r' + \mb{u}_0)^\top X\right) - \beta_1'v_r'\sigma\left((\mb{w}_r' + \mb{u}_1)^\top X\right).
\end{align}

Notice that
\begin{align}
\label{eq:pfLemNDL_beta_u_equality_derive}
\beta_0' \mb{u}_0 + \beta_1' \mb{u}_1 &= \beta_0'(\Delta \ubar{\mb{w}}_r - \Delta \mb{w}_r) + \beta_1'(\Delta \mb{w}_{r+1} - \Delta \mb{w}_r) \nonumber \\ 
&=\frac{1}{\lambda_0'}\Delta \ubar{\mb{w}}_r + \frac{|\lambda_1'|}{\lambda_0'}\Delta \mb{w}_{r+1}-\Delta \mb{w}_r \nonumber \\&= \frac{1}{\lambda_0'}(\Delta \ubar{\mb{w}}_r - \lambda_0'\Delta \mb{w}_r-\lambda_1' \Delta \mb{w}_{r+1} ) \nonumber \\
&= \mb{0}.
\end{align}
where the last equality follows from \eqref{eq:pfThmSL_ND_localmin_auxiliary_perturb}. This implies
\begin{equation}
v_r'\sigma'\left(\ubar{\mb{w}}_r'^\top \mb{x}^{(i)}\right)\left(\beta_0'\mb{u}_0+\beta_1'\mb{u}_1\right)^\top\mb{x}\upbrac{k} = 0.
\end{equation}
where $\mb{x}\upbrac{k}\in \realvec{d}$ denotes the $k$-th training input, i.e., the $k$-th column of $X$, $k\in[n]$. Adding this zero term to the expression \eqref{eq:pfLemNDL_rewrite_e}, we re-write the $k$-th entry of $\mb{e}$ as
\begin{subequations}
\label{eq:pfLemNDL_ek}
\begin{align}
    e_k  =&~  
    v_r'\sigma\left(\mb{w}_r'^\top \mb{x}\upbrac{k}\right) -\beta_0' v_{r}'\sigma\left((\mb{w}_{r}'+\mb{u}_0)^\top \mb{x}\upbrac{k}\right)-\beta_1' v_{r}'\sigma\left((\mb{w}_{r}'+\mb{u}_1)^\top \mb{x}\upbrac{k}\right) \nonumber \\
    &~+ v_r'\sigma'\left(\mb{w}_r'^\top \mb{x}\upbrac{k}\right)\left(\beta_0'\mb{u}_0+\beta_1'\mb{u}_1\right)^\top\mb{x}\upbrac{k}
    \\
    = &~ \sum^1_{j=0}\beta_j' v'_r \left[\sigma\left(\mb{w}_r'^\top \mb{x}\upbrac{k}\right)- \sigma\left((\mb{w}_r' + \mb{u}_j )^\top \mb{x}\upbrac{k}\right) + \sigma'\left(\mb{w}'^\top_r\mb{x}\upbrac{k}\right)\mb{u}_j^\top \mb{x}\upbrac{k} \right] .
\end{align}
\end{subequations}
For $j\in\{0,1\}$ and $k\in[n]$, define $F_{j,k}:[0,1] \rightarrow \mathbb{R} $ as
\begin{equation}
F_{j,k}(t) = \sigma\left(\mb{w}_r'^\top\mb{x}\upbrac{k}\right) - \sigma\left((\mb{w}_r' + t \cdot\mb{u}_j )^\top \mb{x}\upbrac{k}\right)   + t \sigma'\left(\mb{w}_r'^\top\mb{x}\upbrac{k}\right) \mb{u}_j^\top\mb{x}\upbrac{k}.
\end{equation}
This is a twice continuous differentiable function. We can readily verify $F_{j,k}(0) = F'_{j,k}(0) = 0$, and
\begin{equation}
\label{eq:pfLemNDL_Fpp}
    F''_{j,k}(t) = -\sigma''\left((\mb{w}_r' + t \cdot \mb{u}_j)^\top \mb{x}\upbrac{k}\right)\cdot\left( \mb{u}_j^\top \mb{x}\upbrac{k}\right)^2, \quad j\in\{0,1\}, \quad k\in[n].
\end{equation}
We define
\begin{equation}
\label{eq:pfLemNDL_def_big_M}
    M(\delta)\triangleq\max_{\substack{\theta'\in \bar{B}(\theta,\delta)\\j\in\{0,1\}, k\in[n]\\ t\in[0,1]}} \left|\sigma''\left((\mb{w}_r' + t \cdot \mb{u}_j)^\top \mb{x}\upbrac{k}\right)\right|\cdot \left\|\mb{x}\upbrac{k}\right\|_2^2
\end{equation}
where $\bar{B}(\theta,\delta)$ denotes the closure of $B(\theta,\delta)$, which is a compact set. Thus, the maximum in \eqref{eq:pfLemNDL_def_big_M} can be achieved, and $M(\delta)$ is well defined as a finite non-negative value for any $\delta >0$. Also, given a $\delta>0$, $M(\delta)$ is independent of $\Delta \mb{v}$, $\Delta W$, and $t$. Combining \eqref{eq:pfLemNDL_Fpp} and \eqref{eq:pfLemNDL_def_big_M}, we have $|F''_{j,k}(t)| \leq M(\delta)\cdot \|\mb{u}_j\|^2_2$. Then, from \eqref{eq:pfLemNDL_ek} we have
\begin{align}
    |e_k| &=  \left|\sum^1_{j=0}\beta_j' v_r'\int_{0}^1\int_{0}^t F''_{j,k}(\tau)d\tau dt\right| \nonumber \\
    &\leq \sum^1_{j=0}\beta_j' |v_r'| \int_{0}^1\int_{0}^t  \|\mb{u}_j\|^2_2 M(\delta) d\tau dt   \nonumber \\
    &\leq  \frac{1}{2}|v_r'|\cdot M(\delta) \sum^1_{j=0}\beta_j'\|\mb{u}_j\|^2_2.
\end{align}
Note that $\beta_j' < 4\beta_j$. Further, there exists sufficiently small $\delta_{1,1}>0$ such that $|v'_r| \leq 2|v_r|$. Thus, for any $0 < \delta \leq \delta_{1,1}$, we have 
\begin{equation}
    |e_k| \leq 4|v_r|\cdot M(\delta) \left(\beta_0\|\mb{u}_0\|^2_2 + \beta_1\|\mb{u}_1\|^2_2\right), \quad \forall \theta' \in B(\theta,\delta),
\end{equation}
implying
\begin{equation}
\label{eq:pfLemNDL_bound1_cond1}
    \|\mb{e}\|_2 = \sqrt{\sum^n_{k=1}e_k^2}
    \leq 4\sqrt{n}\cdot |v_r|\cdot M(\delta) \left(\beta_0\|\mb{u}_0\|^2_2 + \beta_1\|\mb{u}_1\|^2_2\right), \quad \forall \theta' \in B(\theta, \delta).
\end{equation}

Next, since $\ubar{\bhat{y}}$ is a continuous function with respect to $\theta$. Thus, for any $\epsilon >0$ there exists a sufficiently small $\delta_{1,2}>0$ such that
\begin{equation}
\label{eq:pfLemNDL_bound1_cond2}
    \left\|\ubar{\bhat{y}}'-\ubar{\bhat{y}}\right\|_2 \leq \frac{\epsilon}{8\sqrt{n}\cdot|v_r|\cdot M(1)}\quad  \forall \theta' \in B(\theta,\delta_{1,2}).
\end{equation}
Let $\delta_1 = \min\{\delta_{1,1},\delta_{1,2},1\}$, with \eqref{eq:pfLemNDL_bound1_cond1} and \eqref{eq:pfLemNDL_bound1_cond2} we have that for any $\theta' \in B(\theta,\delta_1)$,
\begin{subequations}
\label{eq:pfLemNDL_ineq1}
\begin{align}
    \left\|(\ubar{\bhat{y}}'-\ubar{\bhat{y}}) + \mb{e}\right\|^2_2  &= \left\|\ubar{\bhat{y}}'-\ubar{\bhat{y}}\right\|^2_2 + 2\mb{e}^\top (\ubar{\bhat{y}}'-\ubar{\bhat{y}}) + \|\mb{e}\|^2_2 \\
    &\geq \left\|\ubar{\bhat{y}}'-\ubar{\bhat{y}}\right\|^2_2 - 2\|\mb{e}\|_2\cdot \left\|\ubar{\bhat{y}}'-\ubar{\bhat{y}}\right\|_2 \\
    \label{eq:pfLemNDL_bound1_1}
    &\geq \left\|\ubar{\bhat{y}}'-\ubar{\bhat{y}}\right\|_2^2 - \frac{M(\delta_{1})}{M(1)}\cdot \epsilon \left(\beta_0\|\mb{u}_0\|^2_2 + \beta_1\|\mb{u}_1\|^2_2\right) \\
    \label{eq:pfLemNDL_bound1_2}
    &\geq \left\|\ubar{\bhat{y}}'-\ubar{\bhat{y}}\right\|_2^2 - \epsilon \left(\beta_0\|\mb{u}_0\|^2_2 + \beta_1\|\mb{u}_1\|^2_2\right)
\end{align}
\end{subequations}
where \eqref{eq:pfLemNDL_bound1_2} follows from the fact that $M$ is monotonically increasing by its definition \eqref{eq:pfLemNDL_def_big_M}. We obtain the first inequality of the lemma.

In the sequel we bound $\langle\Delta \mb{y}, \mb{e}\rangle$. Define a function $G_{a,\mb{b}}: \realvec{d}\rightarrow \mathbb{R}$ parameterized by $a\in\mathbb{R}$ and $\mb{b}\in\mathbb{R}^d$ as
\begin{equation}
    G_{a,\mb{b}}(\mb{u}) = - a \left\langle \Delta \mb{y}, \sigma\left(\sum^d_{k=1}(b_k+u_k) \mb{x}_k\right)\right\rangle
\end{equation}
where $\mb{x}_k \in \mathbb{R}^{n\times 1}$ denotes (the transpose of) the $k$ row of $X$. By \eqref{eq:pfLemNDL_rewrite_e} we have
\begin{align}
\label{eq:pfLemNDL_bound2_cond}
\langle \Delta \mb{y}, \mb{e}\rangle & = \left\langle \Delta \mb{y}, v_r'\sigma(\mb{w}_r'^\top X) - \beta_0'v_r'\sigma\left((\mb{w}_r' + \mb{u}_0)^\top X\right) - \beta_1' v_r'  \sigma\left((\mb{w}_r' + \mb{u}_1)^\top X\right)\right\rangle \\
&=-\sum^1_{j=0}\beta_j'v_r'\left\langle\Delta \mb{y}, \sigma\left(\sum^d_{k=1}(w_{r,k}'+u_{j,k})\mb{x}_k\right)\right\rangle + v_r'\left\langle\Delta \mb{y}, \sigma\left(\sum^d_{k=1}w_{r,k}'\mb{x}_k\right) \right\rangle \\
\label{eq:pfLemNDL_bound2_cond_derive}
&= \beta_0' G_{v_r',\mb{w}_r'}(\mb{u}_0) + \beta_1' G_{v_r',\mb{w}_r'}(\mb{u}_1) - G_{v_r',\mb{w}_r'}(\mb{0})
\end{align}
where $u_{i,k}$ denotes the $k$-th entry of $\mb{u}_i$. 

Now, consider the Hessian matrix of $G_{a,\mb{b}}(\mb{u})$ with respect to $\mb{u}$, denoted by $H_{G}(\mb{u}; a,\mb{b}) \in\realmat{d}{d}$. The entries of $H_{G}(\mb{u}; a,\mb{b})$
\begin{equation}
    \left(H_{G}(\mb{u}; a,\mb{b})\right)_{j,j'} =  -a \left\langle \Delta \mb{y}, \sigma''\left(\sum^d_{k=1}(b_k+u_k)\mb{x}_k\right)\circ \mb{x}_j\circ \mb{x}_{j'}\right\rangle, ~~ j,j'\in[d].
\end{equation}
We have
\begin{subequations}
\begin{align}
    \left(H_{G}(\mb{0}; v_r,\mb{w}_r)\right)_{j,j} &=  -v_r \left\langle \Delta \mb{y}, \sigma''\left(\sum^d_{k=1}w_{r,k}\mb{x}_k\right)\circ \mb{x}_j\circ \mb{x}_{j'}\right\rangle \\
    \label{eq:pfLemNDL_hessian_derive1}
    & = - \lambda_0 \ubar{v}_r \left \langle \Delta \ubar{\mb{y}},
    \sigma''\left(\sum^d_{k=1}\ubar{w}_{r,k}\mb{x}_k\right)\circ \mb{x}_j\circ \mb{x}_{j'}\right\rangle \\
    \label{eq:pfLemNDL_hessian_derive2}
    & = -\lambda_0 \left(\ubarHin_r\right)_{j,j'}
\end{align}    
\end{subequations}
where \eqref{eq:pfLemNDL_hessian_derive1} follows from $v_r = \lambda_0 \ubar{v}_r$, $\mb{w}_r = \ubar{\mb{w}}_r$, and $\Delta \ubar{\mb{y}}$, while \eqref{eq:pfLemNDL_hessian_derive2} is by the definition \eqref{eq:def_inner_Hessian}. Then, by the assumption $\lambda_0 >0$ and $\lambda_{\max}(\ubarHin_r) < 0$, we readily have that $H_{G}(\mb{0}; v_r,\mb{w}_r) = - \lambda_0 \ubarHin_r$ is positive definite. Further, we note that $H_{G}(\mb{u}; a,\mb{b})$ is a continuous function of $(\mb{u},a,\mb{b})$. Thus, there exists a sufficiently small $\delta_{2}>0$ such that for any $\theta' \in B(\theta,\delta_{2})$ and $\mb{u}\in B(\mb{0},2C_2\delta_2)$,\footnote{Recall that the constant $C_2 =2\sqrt{2}\lambda_0$. From \eqref{eq:pfThmSL_ND_localmin_perturbation_range}, we have $\ubar{\theta}' \in B(\ubar{\theta}, C_2\delta_{2})$ as long as $\theta' \in B(\theta,\delta_2)$. This implies $\|\Delta \ubar{\mb{w}}_r\|_2 < C_2\delta_{2}$. Further, by \eqref{eq:pfThmSL_ND_localmin_def_ub} and $C_2>1$, $\|\mb{u}_j\|_2 \leq \sqrt{C_2^2+1}\delta_{2} < 2C_2\delta_{2}$, $j\in\{0,1\}$.} we have
\begin{equation}
    \lambda_{\min}\left(H_G(\mb{u};v'_r, \mb{w}'_r)\right) > \frac{1}{2}\lambda_{\min}\left(H_G(\mb{0};v_r,\mb{w}_r)\right) = -\frac{\lambda_0}{2}\lambda_{\max}(\ubarHin_r) >0.
\end{equation}
Thus, $G_{v_r',\mb{w}_r'}(\mb{u}) + \lambda_0 \cdot \lambda_{\max}(\ubarHin_r)\left\|\mb{u}\right\|^2_2/2$ is convex for all $\mb{u}\in B(\mb{0}, 2C_2\delta_{2})$. By Jensen's inequality, we have
\begin{align}
\label{eq:pfLemNDL_bound2_cond2}
    \sum^1_{j=0}\beta_j'\left[G_{v_r',\mb{w}_r'}(\mb{u}_j)+\frac{\lambda_0}{2}\lambda_{\max}(\ubarHin_r)\|\mb{u}_j\|^2_2\right]
    & \geq G_{v_r',\mb{w}_r'}\left(\sum^1_{j=0}\beta_j'\mb{u}_j\right) + \frac{\lambda_0}{2}\lambda_{\max}(\ubarHin_r)\left\|\sum^1_{j=0}\beta_j'\mb{u}_j\right\|^2_2 \nonumber \\
    %& = G_{v_r',\mb{w}_r'}\left(\sum^2_{i=1}\beta_i'\mb{u}_i\right)
    & = G_{v_r',\mb{w}_r'}\left(\mb{0}\right)
\end{align}
where the last equality follows from \eqref{eq:pfLemNDL_beta_u_equality_derive}. Combining \eqref{eq:pfLemNDL_bound2_cond_derive} and \eqref{eq:pfLemNDL_bound2_cond2}, we have
\begin{equation}
\label{eq:pfLemNDL_ineq2}
    \langle \Delta \mb{y}, \mb{e}\rangle \geq - \frac{\lambda_0}{2}\lambda_{\max}(\ubarHin_r)
    \sum^1_{j=0}\beta_j'\|\mb{u}_j\|^2_2 \geq \frac{1}{8}\left|\lambda_{\max}(\ubarHin_r)\right| \left(\beta_0\|\mb{u}_0\|_2^2 + \beta_1\|\mb{u}_1\|_2^2\right)
\end{equation}
where the second inequality follows from $\beta_j < 4\beta_j'$, $j \in \{0,1\}$ and $\lambda_0>1$. We obtain the second inequality of the lemma.

Finally, setting $\delta = \min\{\delta_1, \delta_2\}$, both \eqref{eq:pfLemNDL_ineq1} and \eqref{eq:pfLemNDL_ineq2} hold. We complete the proof of Lemma \ref{lem:ND_localmin}.

\subsection{Proof of Lemma \ref{lem:saddle_region}}
\label{app:pfLemSR}
\subsubsection{Proof for Positive Eigenvalue}
We first consider the case of $\lambda_{\max}(\ubarHin_r)>0$ and show that $\theta = \neusplit(\ubar{\theta},\pmb{\lambda})$ is a saddle point of \widenet for any $\pmb{\lambda}\in \Lam{+}$. 

Denote the index sets of non-negative and negative $\lambda_j$'s respectively by
\begin{subequations}
\begin{gather}
    J_+ \triangleq \{j \in [0:(m-r)]: \lambda_j \geq 0 \} \\
    J_- \triangleq \{j \in [0:(m-r)]: \lambda_j < 0 \}
\end{gather}
\end{subequations}
and define
\begin{equation}
    S_{+} = \sum_{j\in J_+}\lambda_{j}, ~~~
    S_{-} = \sum_{j\in J_-}\lambda_{j}.
\end{equation}
Because $\sum^{m-r}_{j=0}\lambda_j = 1$, we have $J_+ \not= \varnothing$ and $|S_+| > |S_-|$. The remaining proof is divided into two parts.

\paragraph{Part I: Assume $J_- \not= \varnothing$.} That is, there exists $\lambda_j<0$, implying $S_- <0$. By assumption, $\lambda_{\max}(\ubarHin_r) > 0$, there exists $\mb{a} \in \realvec{d}$ with $\|\mb{a}\|_2 = 1$ such that
\begin{equation}
    \mb{a}^\top \ubarHin_r \mb{a} = \lambda_{\max}(\ubarHin_r)>0.
\end{equation}
We consider a perturbed point $\theta'=(\mb{v}', W') = (\mb{v}+\Delta{\mb{v}}, W+\Delta{W})$ around the embedded point $\theta$. For any $\delta>0$, we set
\begin{subequations}
\label{eq:pfLemSR_pert}
\begin{gather}
    \Delta \mb{w}_i = \mb{0},~~ i\in[r-1], ~~~\Delta \mb{v} = \mb{0}\\
    \Delta \mb{w}_{r+j} = \frac{\alpha}{|S_+|}\mb{a},
    \quad \forall j\in J_+, \quad \Delta \mb{w}_{r+j'} = \frac{\alpha}{|S_-|}\mb{a},
    \quad \forall  j'\in J_-
\end{gather}
\end{subequations}
where $\alpha>0$ is a scalar whose value will be determined later. Clearly, for sufficiently small $\alpha$, $\theta' \in B(\theta,\delta)$. For $i \in [r-1]$, \eqref{eq:pfLemSR_pert} implies $\Delta \mb{t}_i=0$. For $i\in[r:m]$, we perform Taylor expansion on $\Delta \mb{t}_i$ and obtain 
\begin{align}
    \Delta \mb{t}_i^\top
    &= v_i \cdot \sigma((\mb{w}_r + \Delta \mb{w}_i)^\top X) - v_i \cdot \sigma(\mb{w}_r^\top X) \nonumber \\
    &=\lambda_{i-r} \ubar{v}_r \Delta \mb{w}_i^\top \left(\mb{z}_{r}'^\top\circ X\right) + \mb{o}^\top(\|\Delta \mb{w}_{i}\|_2)
\end{align}
where with some abuse of notation we denote
\begin{equation}
\label{eq:pfLemSR_zX}
    \mb{z}_r'^\top\circ X = \begin{bmatrix}
(\mb{z}_r'\circ \mb{x}_1)^\top \\
(\mb{z}_r'\circ \mb{x}_2)^\top \\
\vdots\\
(\mb{z}_r'\circ \mb{x}_d)^\top
\end{bmatrix}
\in \realmat{d}{n}. 
\end{equation}
Then, we have
\begin{align}
\label{eq:pfLemSR_norm}
    \left\|\sum^m_{i=1} \Delta \mb{t}_i\right\|_2 &= \left\|\sum^m_{i=r} \Delta \mb{t}_i\right\|_2 = 
    \left\|\sum^m_{i=r} \left[ \lambda_{i-r}\ubar{v}_r \left(\mb{z}_r'^\top\circ X\right)^\top \Delta \mb{w}_i + \mb{o}(\|\Delta \mb{w}_{i}\|_2)\right]\right\|_2 \nonumber \\
    & = \left\|\ubar{v}_r\alpha\left(\frac{\sum_{j\in J_+} \lambda_{j}}{|S_+|} + \frac{\sum_{j'\in J_-} \lambda_{j'}}{|S_-|}\right)
    \left(\mb{z}_r'^\top\circ X\right)^\top \mb{a} + \mb{o}(\alpha)\right\|_2\nonumber \\
    &=\|\mb{o}(\alpha)\|_2 = o(\alpha).
\end{align}
On the other hand, similarly to \eqref{eq:pfThmSL_decompose_delta_t2} we have
\begin{align}
\label{eq:pfLemSR_decompose_inner_prod}
    \langle \Delta \mb{y}, \Delta \mb{t}_i \rangle & =   \frac{1}{2}  v_i \cdot \sum^{d}_{j=1}\sum^{d}_{j'=1}\Delta w_{i,j} \Delta w_{i,j'}\cdot \langle \Delta \mb{y} , \mb{z}_i''\circ \mb{x}_j \circ \mb{x}_{j'} \rangle + v_i \cdot \langle \Delta \mb{y}, \mb{o}(\|\Delta \mb{w}_i\|^2_2)\rangle \nonumber \\
    & = \frac{1}{2}\Delta \mb{w}_i^\top \Hin_i \Delta\mb{w}_i + o(\|\Delta \mb{w}_i\|^2_2).
\end{align}
Further, noting that $\Hin_{r+j}=\lambda_{j}\ubarHin_r$ for $j\in[0:(m-r)]$, we obtain
\begin{subequations}
\label{eq:pfLemSR_inner_prod}
\begin{align}
    \left \langle \Delta \mb{y}, \sum^m_{i=1} \Delta \mb{t}_i\right \rangle &= \sum_{j \in J_+}\frac{\alpha^2}{2|S_+|^2}\mb{a}^\top \Hin_j \mathbf{a} + \sum_{j' \in J_-}\frac{\alpha^2}{2|S_-|^2}\mb{a}^\top \Hin_{j'} \mathbf{a} +  o(\|\alpha \mb{a}\|^2_2) \\
    & = \frac{\alpha^2}{2}\left(\sum_{j\in J_+}\frac{\lambda_j}{|S_+|^2} + 
    \sum_{j'\in J_-}\frac{\lambda_{j'}}{|S_-|^2}\right) \mb{a}^\top \ubarHin_r \mb{a} + o(\alpha^2) \\
    & = \frac{\alpha^2}{2}\left(\frac{1}{|S_+|} - \frac{1}{|S_-|}\right)\lambda_{\max}(\ubarHin_r) + o(\alpha^2).
\end{align}
\end{subequations}
Combining \eqref{eq:pfLemSR_norm}, \eqref{eq:pfLemSR_inner_prod} and the decomposition \eqref{eq:decompose_empirical_loss2}, for sufficiently small $\alpha$, the difference of the empirical loss is given by
\begin{align}
    L(\theta')-L(\theta) &= \frac{1}{2}\left\|\sum^m_{i=1} \Delta \mb{t}_i\right\|^2_2 + \left\langle \Delta \mb{y}, \sum^m_{i'=1} \Delta \mb{t}_{i'} \right\rangle \nonumber\\
    & = \frac{\alpha^2}{2}\left(\frac{1}{|S_+|} - \frac{1}{|S_-|}\right)\lambda_{\max}(\ubarHin_r) + o(\alpha^2) \nonumber \\
    &< 0
\end{align}
where the last inequality holds due to $|S_+| > |S_-|$. Therefore, we can find a perturbed point with smaller empirical loss in an arbitrarily small neighborhood of $\theta$, and hence $\theta$ is not a local minimum. From \eqref{lem:no_localmax}, $\theta$ is not a local maximum either. Therefore, $\theta$ is a saddle point of \widenet.

\paragraph{Part II: Assume $J_- = \varnothing$.} Because $\pmb{\lambda} \in \Lam{+}$, there exists $j_0\in[0,(m-r)]$ with $\lambda_{j_0} \leq 0$. Therefore, we must have $\lambda_{j_0} = 0$. Without loss of generality we assume $j_0=0$. By the definition of inner Hessian in \eqref{eq:def_inner_Hessian}, $\lambda_{\max}(\ubarHin_r)>0$ implies $\ubar{v}_r \not=0$. Given an arbitrary $\delta>0$, we construct a perturbed point $\theta'$ as follows
\begin{subequations}
\begin{gather}
    W' = W, \quad v_i' = v_i, ~~ i\in [r-1]\\
    v_{r+j}' = \lambda_{j}' \ubar{v}_r, ~~ j\in[0:(m-r)]
\end{gather}
\end{subequations}
where $\{\lambda_j'\}^{m-r}_{j=0}$ are defined as
\begin{equation}
\lambda_0' = \lambda_0  -\frac{\sqrt{2}\gamma}{2|\ubar{v}_r|},\quad  \lambda_1' = \lambda_1  +\frac{\sqrt{2}\gamma}{2|\ubar{v}_r|} , \quad \lambda_{j}' = \lambda_{j},\quad  j\in[2:(m-r)] 
\end{equation}
with $\gamma>0$. We readily have $ \theta'\in B(\theta,\delta/2)$ if $\gamma<\delta/2$. Further, because $\sum^r_{i=0}\lambda_i' = 1$, $\theta'$ is generated by splitting the $r$-th neuron of \narrownet at $\ubar{\theta}$ with coefficients $\pmb{\lambda}' = [\lambda_0',\cdots, \lambda_{m-r}']^\top$. By Proposition \ref{prop:stat_preserve}, $\theta'$ is a stationary point of \widenet which admits the same empirical loss with $\theta$. Further, noting that $\lambda_0=0$, we have $\lambda_0'<0$. By the analysis in Part I, $\theta'$ is a saddle point. Thus, there exists $\theta'' \in B(\theta',\delta/2)$ such that
\begin{equation}
L(\theta'') < L(\theta') = L(\theta).
\end{equation}
Since $\theta'' \in B(\theta,\delta)$ and $\delta$ can be arbitrarily small, we have that $\theta$ is a saddle point of \widenet.

Combining Part I and Part II, we conclude that if $\lambda_{\max}(\ubarHin_r)>0$, $\theta$ is a saddle point of \widenet for any $\pmb{\lambda} \in \Lam{+}$. We obtain the first conclusion of the lemma.

\subsubsection{Proof for Negative Eigenvalue}
We next consider the case of $\lambda_{\min}(\ubarHin_r)<0$ and show that $\theta = \neusplit(\ubar{\theta},\pmb{\lambda})$ is a saddle point of \widenet for any $\pmb{\lambda} \in \Lambda_-$. Without loss of generality we assume $\lambda_j$'s are in descending order, i.e., $\lambda_0\geq \lambda_1\geq \cdots \geq \lambda_{m-r}$,
By $\pmb{\lambda}\in\Lam{-}$, we have $\lambda_0, \lambda_1 \geq 0$. Because $\sum^{m-r}_{j=0}\lambda_j = 1$, we must have $\lambda_0 > 0$. We consider the cases with $\lambda_1>0$ and $\lambda_1 =0$ respectively and divide the remaining proof into two parts.

\paragraph{Part I: Assume $\lambda_1>0$.} 
Because $\lambda_{\min}(\ubarHin_r)<0$, there exists $\mathbf{b} \in \realvec{d}$ with $\|\mb{b}\| =1$ such that 
\begin{equation}
    \mb{b}^\top \ubarHin_r \mb{b} = \lambda_{\min}(\ubarHin_r)<0.
\end{equation}
We consider a perturbed point $\theta'=(\mb{v}', W') = (\mb{v}+\Delta{\mb{v}}, W+\Delta{W})$ around the embedded point $\theta$. For an arbitrary $\delta>0$, we set
\begin{subequations}
\label{eq:pfLemSR_pert2}
\begin{gather}
\label{eq:pfLemSR_pert2_1}
   \Delta \mb{v} = \mb{0},\quad 
   \Delta \mb{w}_i = \mb{0},~~ i\in[r-1]\cup[(r+2):m] \\
\label{eq:pfLemSR_pert2_2}
 \Delta \mb{w}_r = \frac{\beta}{\lambda_0} \mb{b}, \quad \Delta \mb{w}_{r+1} = -\frac{\beta}{\lambda_{1}}\mb{b}
\end{gather}
\end{subequations}
where $\beta>0$ is a scalar whose value will be determined later. Clearly, $\theta'\in B(\theta,\delta)$ if $\beta$ is sufficiently small. For $i\in[r-1]\cup [(r+2):m]$, \eqref{eq:pfLemSR_pert2} implies $\Delta \mb{t}_i = \mb{0}$. For $i=r, r+1$, we perform Taylor expansion on $\Delta \mb{t}_i$ and obtain 
\begin{align}
\label{eq:pfLemSR_delta_t_2}
    \Delta \mb{t}_i^\top
    &= v_i \cdot \sigma((\mb{w}_r + \Delta \mb{w}_i)^\top X) - v_i \cdot \sigma(\mb{w}_r^\top X) \nonumber \\
    &=\lambda_{i-r} \ubar{v}_r \Delta \mb{w}_i^\top \left(\mb{z}_{r}'^\top\circ X\right) + \mb{o}^\top(\|\Delta \mb{w}_i\|_2)
\end{align}
where $\mb{z}_{r}'^\top\circ X$ is given by \eqref{eq:pfLemSR_zX}. Then, we have
\begin{align}
\label{eq:pfLemSR_norm_2}
    \left\|\sum^m_{i=1} \Delta \mb{t}_i\right\|_2 &= \left\|\sum^{r+1}_{i=r} \Delta \mb{t}_i\right\|_2 = 
    \left\|\sum^{r+1}_{i=r} \lambda_{i-r}\ubar{v}_r \left(\mb{z}_r'^\top\circ X\right)^\top \Delta \mb{w}_i +\sum^{r+1}_{j=r} \mb{o}(\|\Delta \mb{w}_{i}\|_2)\right\|_2 \nonumber \\
    & = \left\|\ubar{v}_r(\beta-\beta)
    \left(\mb{z}_r'^\top\circ X\right)^\top \mb{b} + \mb{o}(\beta)\right\|_2\nonumber \\
    &=\|\mb{o}(\beta)\|_2 = o(\beta).
\end{align}
On the other hand, similarly to \eqref{eq:pfThmSL_decompose_delta_t2} we have
\begin{align}
    \langle \Delta \mb{y}, \Delta \mb{t}_i \rangle & =   \frac{1}{2}  v_i \cdot \sum^{d}_{j=1}\sum^{d}_{j'=1}\Delta w_{i,j} \Delta w_{i,j'}\cdot \langle \Delta \mb{y} , \mb{z}_i''\circ \mb{x}_j \circ \mb{x}_{j'} \rangle + v_i \cdot \langle \Delta \mb{y}, \mb{o}(\|\Delta \mb{w}_i\|^2_2)\rangle \nonumber \\
    & = \frac{1}{2}\Delta \mb{w}_i^\top \Hin_i \Delta\mb{w}_i + o(\|\Delta \mb{w}_i\|^2_2).
\end{align}
Further, noting that $\Hin_{r+j}=\lambda_{j}\ubarHin_r$ for $j\in\{0,1\}$, we obtain
\begin{subequations}
\label{eq:pfLemSR_inner_prod_2}
\begin{align}
    \left \langle \Delta \mb{y}, \sum^m_{i=1} \Delta \mb{t}_i\right \rangle &= \frac{\lambda_0}{2}\Delta \mb{w}_r^\top \ubarHin_r \mb{w}_r +  \frac{\lambda_1}{2}\Delta \mb{w}_{r+1}^\top \ubarHin_r \mb{w}_{r+1} + o(\|\Delta \mb{w}_r\|^2_2) + o(\|\Delta \mb{w}_{r+1}\|^2_2) \\ 
    \label{eq:pfLemSR_inner_prod_2_1}
    & = \left(\frac{\beta^2}{2\lambda_0} + \frac{\beta^2}{2\lambda_1}\right)\mb{b}^\top \ubarHin_r \mathbf{b}  +  o(\|\beta \mb{b}\|^2_2) \\
    \label{eq:pfLemSR_inner_prod_2_2}
    & =  \frac{\beta^2}{2} \left(\frac{1}{\lambda_0}+\frac{1}{\lambda_1}\right) \lambda_{\min}(\ubarHin_r) + o(\beta^2)
\end{align}
\end{subequations}
where \eqref{eq:pfLemSR_inner_prod_2_1} follows from \eqref{eq:pfLemSR_pert2}.
Finally, combining \eqref{eq:pfLemSR_norm_2}, \eqref{eq:pfLemSR_inner_prod_2}, and the decomposition \eqref{eq:decompose_empirical_loss2}, the difference of the empirical loss is given by
\begin{align}
    L(\theta')-L(\theta) &= \frac{1}{2}\left\|\sum^m_{i=1} \Delta \mb{t}_i\right\|^2_2 + \left\langle \Delta \mb{y}, \sum^m_{i=1} \Delta \mb{t}_i \right\rangle \nonumber\\
    & = \frac{\beta^2}{2}\left(\frac{1}{\lambda_0} + \frac{1}{\lambda_1}\right)\lambda_{\min}(\ubarHin_r) + o(\beta^2) \nonumber \\
    &< 0
\end{align}
for sufficiently small $\beta$, where the last inequality holds due to $\lambda_{\min}(\ubarHin_r) <0$ and $\lambda_0,\lambda_1>0$ by assumption. Therefore, we can find a perturbed point with a smaller empirical loss in an arbitrarily small neighborhood of $\theta$, and hence $\theta$ is not a local minimum. From Lemma \ref{lem:no_localmax}, $\theta$ is not a local maximum either. Therefore, $\theta$ is a saddle point of \widenet.

\paragraph{Part II: Assume $\lambda_1 =0$.} By the definition of inner Hessian in \eqref{eq:def_inner_Hessian}, $\lambda_{\min}(\ubarHin_r)<0$ implies $\ubar{v}_r \not=0$. Given an arbitrary $\delta>0$, we construct a perturbed point $\theta'$ as follows
\begin{subequations}
\begin{gather}
    W' = W, \quad v_i' = v_i, ~~ i\in [r-1]\\
    v_{r+j}' = \lambda_{j}' \ubar{v}_r, ~~ j\in[0:(m-r)]
\end{gather}
\end{subequations}
where $\{\lambda_j'\}^{m-r}_{j=0}$ are defined as
\begin{equation}
\lambda_0' = \lambda_0  -\frac{\sqrt{2}\gamma}{2|\ubar{v}_r|},\quad  \lambda_1' = \lambda_1  +\frac{\sqrt{2}\gamma}{2|\ubar{v}_r|} , \quad \lambda_{j}' = \lambda_{j},\quad  j\in[2:(m-r)] 
\end{equation}
with $0<\gamma<\sqrt{2}|\ubar{v}_r|\lambda_0$. We readily have $\theta'\in B(\theta,\delta/2)$ if $\gamma<\delta/2$. Further, because $\sum^r_{i=0}\lambda_i' = 1$, $\theta'$ is generated by splitting the $r$-th neuron of \narrownet at $\ubar{\theta}$ with coefficients $\pmb{\lambda}' = [\lambda_0',\cdots, \lambda_{m-r}']^\top$. By Proposition \ref{prop:stat_preserve}, $\theta'$ is a stationary point of \widenet which admits the same empirical loss with $\theta$. Further, noting that $\lambda_0>\lambda_1=0$, we have $\lambda_0'>\lambda_1'>0$. By the analysis in Part I, $\theta'$ is a saddle point. Thus, there exists $\theta'' \in B(\theta',\delta/2)$ such that
\begin{equation}
L(\theta'') < L(\theta') = L(\theta).
\end{equation}
Since $\theta'' \in B(\theta,\delta)$ and $\delta$ can be arbitrarily small, we have that $\theta$ is a saddle point of \widenet.

Combining Part I and Part II, we conclude that if $\lambda_{\min}(\ubarHin_r)<0$, $\theta$ is a saddle point of \widenet for any $\pmb{\lambda} \in \Lam{-}$. We obtain the second conclusion of the lemma.

\section{Connection between Hessian Matrix and Inner Hessian}
\label{app:inner_hessian}

Note that by \eqref{eq:pfThmSL_stationary_cond_w}, the partial gradient of $L$ with respect to $\mb{w}_i$ is

\begin{equation}
    \frac{\partial L}{\partial \mb{w}_i} =
    v_i \begin{bmatrix}
    \langle \Delta \mb{y}, \mb{z}_i' \circ \mb{x}_1 \rangle \\
    \langle \Delta \mb{y}, \mb{z}_i' \circ \mb{x}_2 \rangle \\
    \vdots \\
    \langle \Delta \mb{y}, \mb{z}_i' \circ \mb{x}_d \rangle 
    \end{bmatrix}.
\end{equation}

Thus, for $1\leq j\leq m$, the $j$-th entry of the partial gradient with respect to $\mb{w}_i$ is

\begin{align}
    \left(\frac{\partial L}{\partial \mb{w}_i}\right)_j =& \langle \hat{\mb{y}} -  \mb{y}, \mb{z}_i' \circ \mb{x}_j \rangle \nonumber\\
    =& \sum_{k=1}^n \left[v_i\cdot\sigma\left(\sum_{l=1}^d w_{i,l}\cdot X_{l,k}\right)-y_k\right]\cdot \left[v_i\cdot\sigma'\left(\sum_{l=1}^d w_{i,l}\cdot X_{l,k}\right)\cdot X_{j,k}\right].
\end{align}

Taking a further derivative step, we have that for $1\leq j, j'\leq d$, 

\begin{align}\label{eq::hessian_entrywise}
    \left(\frac{\partial^2 L}{\partial \mb{w}_i^2}\right)_{j, j'} = & \sum_{k=1}^n\left[v_i\cdot\sigma'\left(\sum_{l=1}^d w_{i,l}\cdot X_{l,k}\right)\cdot X_{j',k}\right] \cdot \left[v_i\cdot\sigma'\left(\sum_{l=1}^d w_{i,l}\cdot X_{l,k}\right)\cdot X_{j,k}\right]\nonumber\\
    & +\sum_{k=1}^n \left[v_i\cdot\sigma\left(\sum_{l=1}^d w_{i,l}\cdot X_{l,k}\right)-y_k\right]\cdot \left[v_i\cdot\sigma''\left(\sum_{l=1}^d w_{i,l}\cdot X_{l,k}\right)\cdot X_{j,k}X_{j',k}\right].
\end{align}

Note that for any $1\leq k\leq n$, 
\begin{align}
    \hat{y}_k = & v_i\cdot\sigma\left(\sum_{l=1}^d w_{i,l}\cdot X_{l,k}\right), \nonumber\\ 
    \left(\frac{\partial\hat{y}_k}{\partial \mb{w}_i}\right)_j = & v_i\cdot\sigma'\left(\sum_{l=1}^d w_{i,l}\cdot X_{l,k}\right)\cdot X_{j,k}, \nonumber\\
    z''_{i,k} = & \sigma''\left(\sum_{l=1}^d w_{i,l}\cdot X_{l,k}\right),
\end{align}
we can rewrite \eqref{eq::hessian_entrywise} into a more compact form:
\begin{align}\label{eq:partial_Hessian_decompose_entrywise}
    \left(\frac{\partial^2 L}{\partial \mb{w}_i^2}\right)_{j, j'} = & \sum_{k=1}^n \left(\frac{\partial\hat{y}_k}{\partial \mb{w}_i}\right)_j\cdot \left(\frac{\partial\hat{y}_k}{\partial \mb{w}_i}\right)_{j'} + \sum_{k=1}^n\left(\hat{y}_k - y_k\right)\cdot \left(v_i\cdot z''_{i,k}\cdot X_{j,k}X_{j',k}\right) \nonumber \\
    = &\left\langle J_j(\hat{\mb{y}}, \mb{w}_i), J_{j'}(\hat{\mb{y}}, \mb{w}_i)\right\rangle + \sum^n_{k=1} (\hat{y}_k-y_k) H_{j,j'}(\hat{y}_k, \mb{w}_i),
\end{align}
where $J_j(\hat{\mb{y}}, \mb{w}_i)$ denotes the $j$-th row of $J_{\bhat{y}}^\top(\mb{w}_i)$, which is the Jacobian matrix of $\bhat{y}$ with respect to $\mb{w}_i$, and $H_{j,j'}(\hat{y}_k, \mb{w}_i)$ denotes the $(j, j')$-th entry of $H_{\hat{y}_k}(\mb{w}_i)$, which is the Hessian matrix of $\hat{y}_k$ with respect to $\mb{w}_i$. 

Noting that \eqref{eq:partial_Hessian_decompose_entrywise} is the element-wise representation of \eqref{eq:partial_Hessian_decompose}, we thus obtain the Hessian decomposition \eqref{eq:partial_Hessian_decompose}. Furthermore, the second term of \eqref{eq:partial_Hessian_decompose_entrywise} has a compact representation:
\begin{equation}
    \sum_{k=1}^n\left(\hat{y}_k - y_k\right)\cdot \left(v_i\cdot z''_{i,k}\cdot X_{j,k}X_{j',k}\right) = v_i\langle \bhat{y} -\mb{y}, \mb{z}_i'' \circ \mb{x}_j \circ \mb{x}_{j'} \rangle,
\end{equation}
which is exactly the inner Hessian $\Hin_i$ by the definition in \eqref{eq:def_inner_Hessian}. Therefore, we have verified that $\Hin_i$ is the second term of the Hessian decomposition \eqref{eq:partial_Hessian_decompose}.

\end{document}